\documentclass[runningheads]{llncs}
\usepackage{graphicx}
\usepackage{amsmath,amssymb} 
\usepackage{color}
\usepackage[width=122mm,left=12mm,paperwidth=146mm,height=193mm,top=12mm,paperheight=217mm]{geometry}

\DeclareMathOperator*{\argmin}{arg\,min}  
\DeclareMathOperator*{\argmax}{arg\,max}  

\usepackage[pagebackref]{hyperref}
\hypersetup
{
	colorlinks=false,
	linkbordercolor={0.6 1 1},
	pagebackref=true
	
}
\usepackage{authblk} 
\usepackage{subfiles}
\begin{document}
\pagestyle{headings}
\mainmatter
\def\ECCV16SubNumber{***}  

\title{LiDAR-Camera Calibration using 3D-3D Point correspondences} 

\titlerunning{LiDAR-Camera Calibration using 3D-3D Point correspondences}

\authorrunning{A. Dhall, K. Chelani, V. Radhakrishnan, K.M. Krishna}

\author{Ankit Dhall$^1$, Kunal Chelani$^2$, Vishnu Radhakrishnan$^3$, K. Madhava Krishna$^4$}

\institute{$^1$Vellore Institute of Technology, Chennai \\
$^2$Birla Institute of Technology and Science, Hyderabad \\
$^3$Veermata Jijabai Technological Institute, Mumbai  \\
$^4$International Institute of Information and Technology, Hyderabad
}

\maketitle

\begin{figure}[h]
\centering
\includegraphics[width=\linewidth]{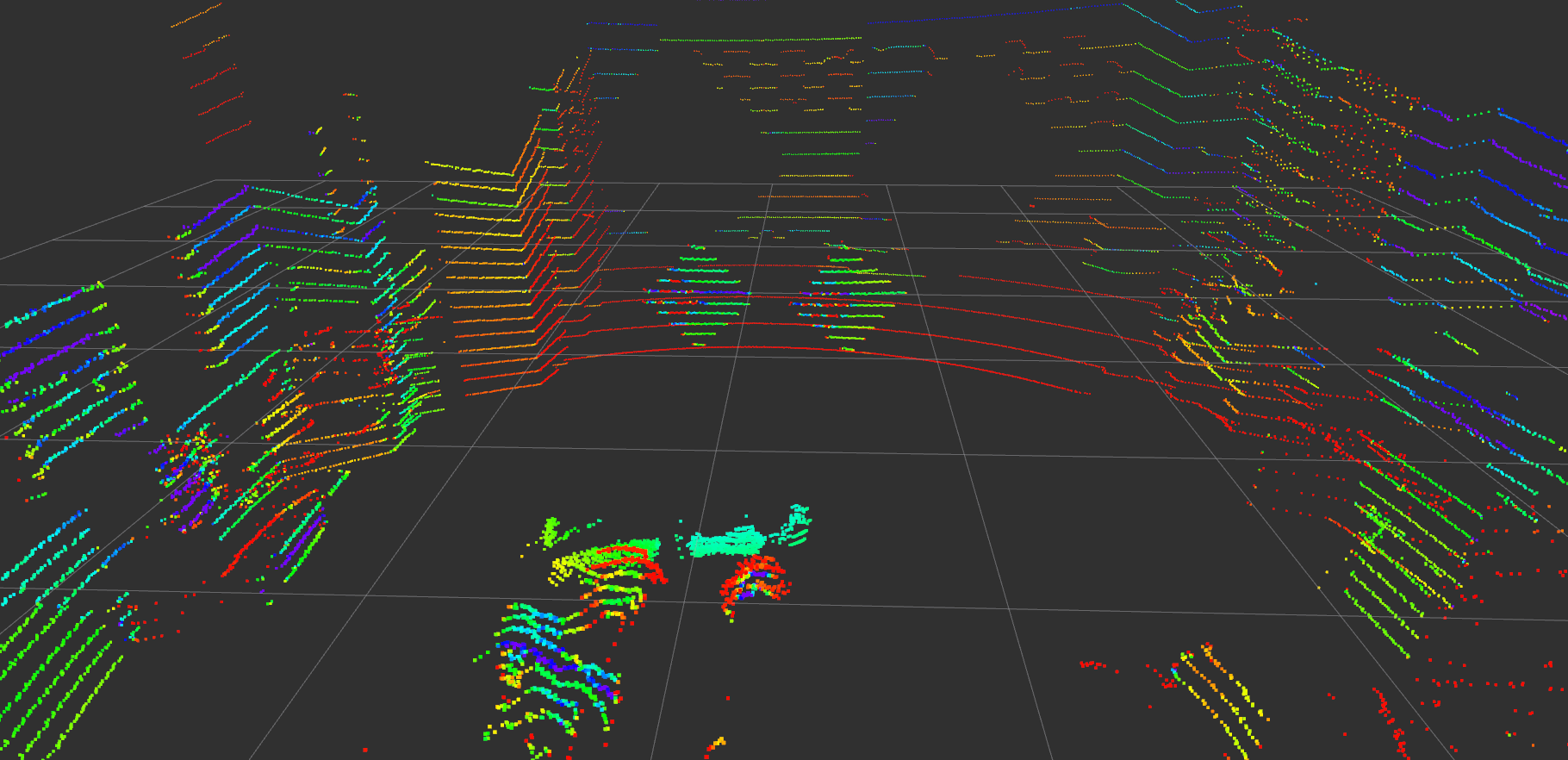}
\caption{LiDAR scan of the experimental setup. 3D point correspondences are obtained in the LiDAR as well as camera frame. The two sets of 3D points are used to solve for a rotation and then a translation.}
\label{fig:LiDAR_Pointcloud_for_20_points}
\end{figure}

\let\thefootnote\relax\footnotetext{Work was done during an internship at Robotics Research Center at IIIT-H}

\begin{abstract}
With the advent of autonomous vehicles, LiDAR and cameras have become an indispensable combination of sensors. They both provide rich and complementary data which can be used by various algorithms and machine learning to sense and make vital inferences about the surroundings. We propose a novel pipeline and experimental setup to find accurate rigid-body transformation for extrinsically calibrating a LiDAR and a camera. The pipeling uses 3D-3D point correspondences in LiDAR and camera frame and gives a closed form solution. We further show the accuracy of the estimate by fusing point clouds from two stereo cameras which align perfectly with the rotation and translation estimated by our method, confirming the accuracy of our method's estimates both mathematically and visually. Taking our idea of extrinsic LiDAR-camera calibration forward, we demonstrate how two cameras with no overlapping field-of-view can also be calibrated extrinsically using 3D point correspondences. The code has been made available as open-source software in the form of a ROS package.

\end{abstract}

\section{Introduction}

Robotic platforms, both autonomous and remote controlled, use multiple sensors such as IMUs, multiple cameras and range sensors. Each sensor provides data in a complementary modality. For instance, cameras provide rich color and feature information which can be used by state-of-the-art algorithms to detect objects of interest (pedestrians, cars, trees, etc.). Range sensors have gained a lot of popularity recently despite being more expensive and also contain moving parts. These can provide rich structural information and if correspondence can be drawn between the camera and the LiDAR, when a pedestrian is detected in an image, it's exact 3D location can be estimated and be used by an autonomous car to avoid obstacles and prevent accidents.

\par
Multiple sensors are employed to provide redundant information which reduces the chance of having erroneous measurements. In the above cases, it is essential to obtain data from  various sensors with respect to a single frame of reference so that data can be fused and redundancy can be leveraged. Marker based\cite{but_velodyne} as well as automatic calibration for LiDAR and cameras has been proposed but methods and experiments discussed in these use the high-density, more expensive LiDAR and do not extend very well when a lower-density LiDAR, such as the VLP-16 is used.

\par
We propose a very accurate and repeatable method to estimate extrinsic calibration parameters in the form of 6 degrees-of-freedom between a camera and a LiDAR.

\section{Sensors and General Setup}
The method we propose makes use of sensor data from a LiDAR and a camera. The intrinsic parameters of the camera should be known before starting the LiDAR-camera calibration process.

\par
The camera can only sense the environment directly in front of the lens unlike a LiDAR such as the Velodyne VLP-16, which has a 360-degree view of the scene, and the camera always faces the markers directly. Each time data was collected, the LiDAR and camera were kept at arbitrary distance in 3D space. The transformation between them was measured manually. Although, the tape measurement is crude, it serves as a sanity check for values obtained using various algorithms. Measuring translation is easier than rotation. When the rotations were minimal, we assumed them to be zero, in other instances, when there was considerable rotation in the orientation of the sensors, we measured distances and estimated the angles roughly using trigonometry.

\section{Using 2D-3D correspondences} \label{2d_3d}
Before working on our method that uses 3D-3D point correspondences, we tried methods that involved 2D-3D correspondences. We designed our own experimental setup to help calibrate a LiDAR and camera, first, using 2D-3D methods.

\par
The setup involves markers of a specific type: hollow rectangular cardboards. Even normal cardboards work fine, however, as we shall see in the upcoming discussion, provide less correspondences as opposed to a hollowed out rectangular cardboard.  

\begin{figure}[h]
\centering
\includegraphics[width=\linewidth]{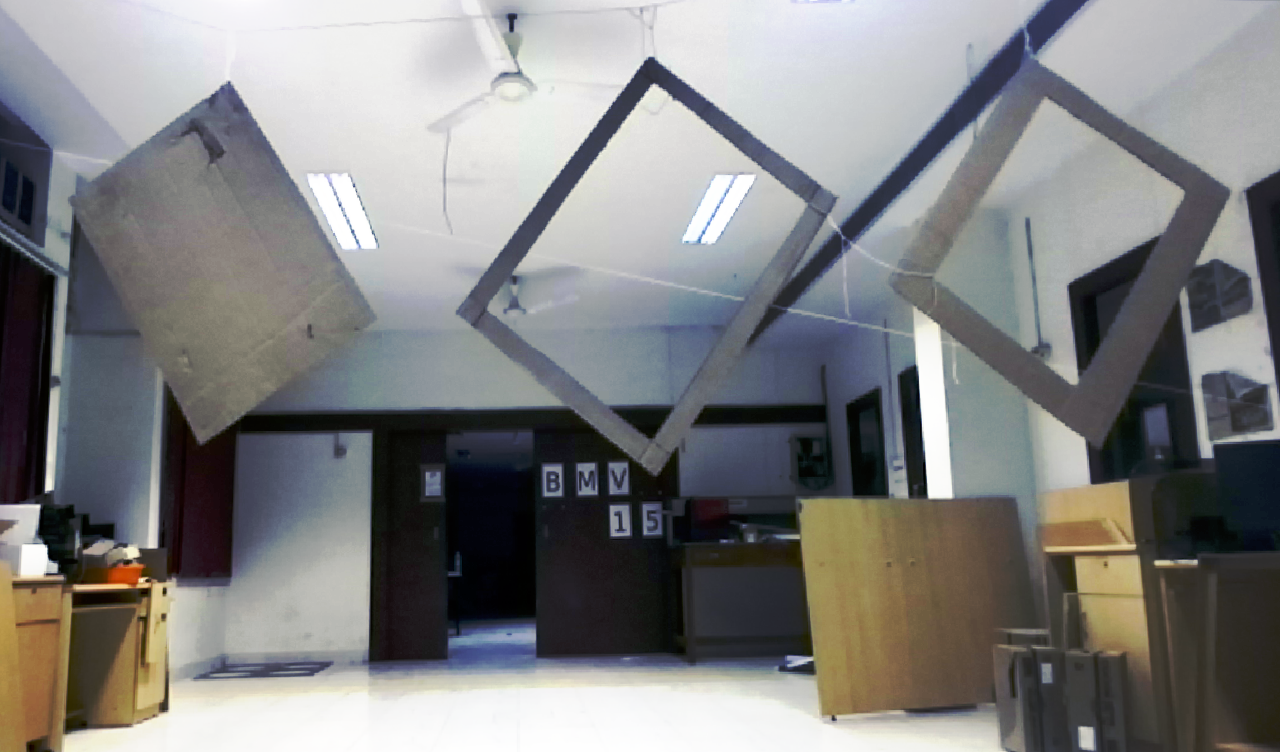}
\caption{Experimental setup with rectangular cardboard cutouts using 2D-3D correspondences.}
\label{fig:rect_20pts_setup}
\end{figure}

\par
This method involves finding the 6-DoF between the camera and the LiDAR by the means of matching 2D-3D point correspondences. 2D correspondences can be easily obtained by manually marking feature points in an image with an accuracy of 3-4 pixels. Obtaining corresponding 3D points is not that straight-forward. For one reason LiDARs does not give a high density point cloud and with increasing distance (away from the LiDAR center) the point cloud becomes more and more sparse.

\par
A planar cardboard can provide 4 corner points i.e. 4 point correspondences. In 3D these points are obtained by line-fitting followed by line-intersection and their 2D correspondences can be obtained by marking pixel co-ordinates. If a hollowed out rectangular cardboard is used, it provides 8 3D-2D point correspondences: 4 corners on the outer rectangle and 4 corners on the inner rectangle; doubling the correspondences, allowing for more data points with lesser number of boards. Such a setup allows to have enough data to run a RaNSaC version of PnP algorithms and also will help reduce noisy data, in general.

\par
We use rectangular (planar cardboard) markers. If in the experimental setup, the markers are kept with one of their sides parallel to the ground, due to the horizontal nature of the LiDAR's scan lines one can obtain the vertical edges, but not necessarily the horizontal ones. To overcome this, we tilt the board to make approximately 45 degrees between one of the edges and the ground plane. With such a setup we always obtain points on all four edges of the board. RanSaC is used to fit lines on the points from the LiDAR.

\begin{figure}[h]
\centering
\includegraphics[width=\linewidth]{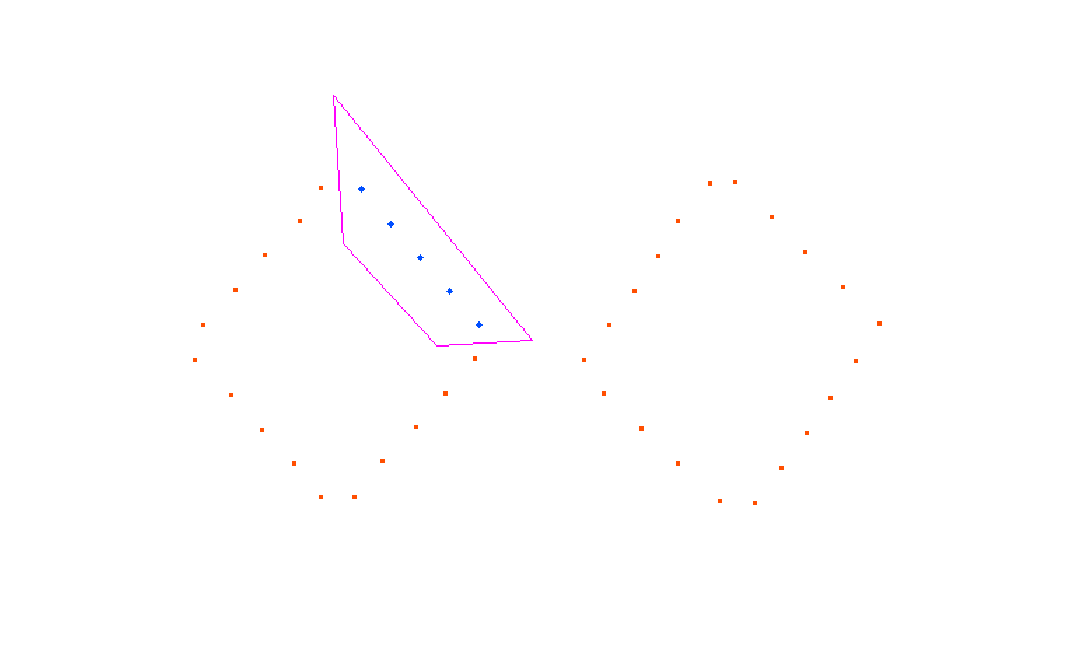}
\caption{Marking line segments in the 3D-pointcloud. The ROS node allows for manually marking segments by drawing polygons around each line segment and also calculate their intersections.}
\label{fig:line_marking}
\end{figure}

\par
The most prominent feature on the marker is the corner. It can be marked with relative easy on the image and since we have quite accurate line equations for the four edges, their intersection is calculated in 3D. Again, these lines may not actually intersect, but come very close. We approximate the corner to the midpoint on the shortest line-segment between the two lines.

\par
As, a check, that this point is indeed a very close approximation to the actual corner, we calculated the length of the shortest line-segment. Also, since we know the dimensions of the cardboard marker, the length of the opposite sides should be very close to each other and also to the actual length measure by tape. We consistently observed that the distance between two line-segments was of the order $10^{-4}$ meters and the error between the edge lengths was about 1 centimeter on average.

\par
Collecting data over multiple experiments, we observed that the edges are extremely close and the corner and intersection are at most off by 0.68 mm on average. An average absolute deviation of 1cm is observed between the expected and estimated edge lengths of the cardboard markers. With the above two observations one can conclude that the intersections are indeed a very accurate approximation of the corner in 3D.
\begin{equation} \label{eq:camera_rt}
\begin{bmatrix}
    u \\
    v \\
    1 \\
\end{bmatrix}
=
\begin{bmatrix}
    f_{x} & \gamma & c_{x} \\
    0     & f_{x}  & c_{y} \\
    0     & 0      & 1
\end{bmatrix}
*
\begin{bmatrix}
    r_{11} & r_{12} & r_{13} & t_{1} \\
    r_{21} & r_{22} & r_{23} & t_{2} \\
    r_{31} & r_{32} & r_{33} & t_{3}
\end{bmatrix}
*
\begin{bmatrix}
    x \\
    y \\
    z \\
    1 \\
\end{bmatrix}
\end{equation}
\par
Using hollowed out markers, we obtained 20 corner points: 2 hollow rectangular markers (8+8 points) and one solid rectangular marker (4 points) increasing the number of point correspondences from our initial experiments.

\par
Perspective n-Point (PnP) finds the rigid-body transformation between a set of 2D-3D correspondences. Equation \ref{eq:camera_rt} shows how the 3D points are projected after applying the $[R|t]$ which is estimated by PnP. Equation \ref{eq:bp_error} represents the general cost-function to solve such a problem.

\begin{equation} \label{eq:bp_error}
\argmin_{R \in SO(3), t \in \mathbb{R}^{3}} ||P(RX+t)-x||^{2}
\end{equation}

where,
\begin{center}
$P$ is the projection operation from 3D to 2D on the image plane \\
$X$ represents points in 3D \\
$x$ represents points in 2D
\end{center}

\par
To begin with, we started with PnP and E-PnP\cite{epnp}. The algorithms seemed to minimize the error; and with manually filtering (refer to table \ref{tab:manual_ransac}) the points (by visualizing the outliers) we were able to lower the back-projection error to 1.88 pixels on average. However, one did not observe the $[R|t]$ close to the values measure by measuring tape between the camera and LiDAR.

\begin{figure}[h]
\centering
\includegraphics[width=0.9\linewidth]{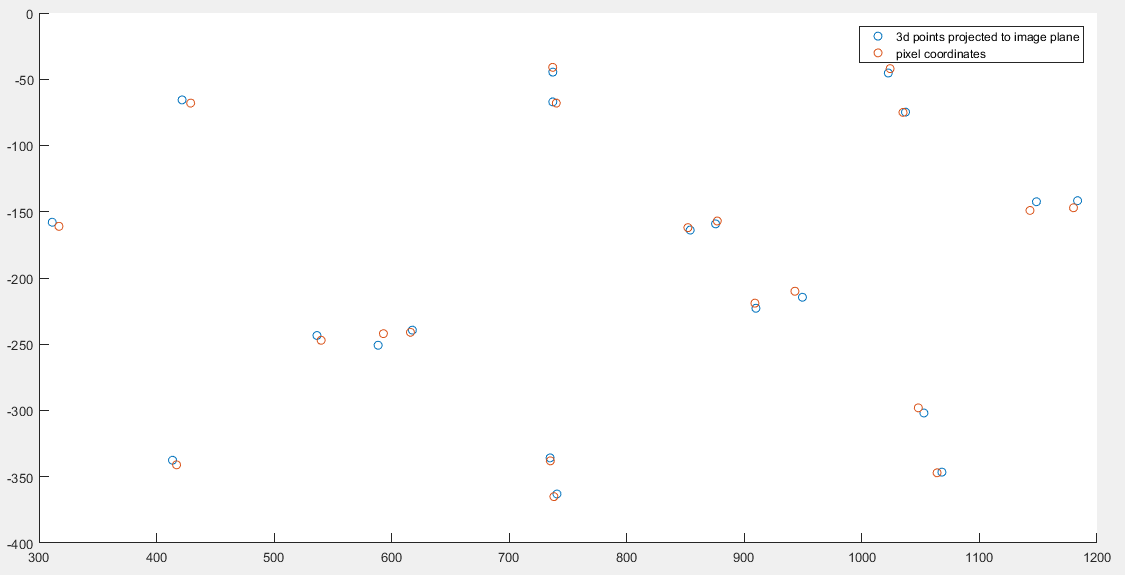}
\caption{Back-projected 3D points and 2D points on the image plane (in pixels) using E-PnP. The 20 points (refer to table \ref{tab:manual_ransac}) seem to be aligned, however, the transformation obtained deviates from the measured values.}
\label{fig:20points_backproj}
\end{figure}

\begin{table}[!htbp]
\centering
\begin{tabular}{|c|c|c|c|c|}
\hline
 & Expected Value & \multicolumn{3}{c}{\# of point correspondences}\\
 & (by tape) & 20 points & 11 points & 8 points \\
\hline
X (in m) & 0 & 0.0094 & 0.0356 & 0.0776 \\
Y (in m) & -0.51 & -0.6174 & -0.6142 & -0.5348 \\
Z (in m) & -0.755 & -0.7101 & -0.7420 & -0.7447 \\ \hline
Rot x (in deg) & - & -5.4194 & -5.4547 & -3.5616 \\
Rot y (in deg) & - & -0.7949 & -1.6225 & -2.5203 \\
Rot z (in deg) & - & -0.5163 & -0.5538 & -0.7039 \\ \hline
Back projection error (in pixels) & - & 4.8211 & 2.9476 & 1.8836 \\
\hline
\end{tabular}
\caption{Comparison of manually measured $[R|t]$ and $[R|t]$ from E-PnP. We tried E-PnP with different number of points: 20, 11 and 8. To eliminate points, we visualized the alignment of the 3D points projected on to the image plane and the corresponding 2D points and manually removed correspondences that were contributing more to the error. Further, we implemented a RaNSaC version of the PnP to do this automatically.}
\label{tab:manual_ransac}
\vspace{-0.7cm}
\end{table}

\par
In a previous experiment, when the LiDAR and camera were quite close (12cm apart) we ran the E-PnP with 12 points and did not obtain the expected values. We observed that we got an error of 10cm and if our expected value is around that measure of granularity then we can expect to obtain noisy values. In subsequent experiments the the camera and LiDAR were kept even farther apart so that the influence of any error is mitigated.

\par
While examining the data, we found that there were some noisy data points who were contributing to a large back projection error. We ran a modified E-PnP with a RanSaC algorithm on top. This would in theory ensure that noisy data is not considered while calculating the rigid body transformation between the camera and the LIDAR. RanSaC selects a random subset of the data, fits a model, estimates data points that are inliers to the fitted model (given a threshold $\epsilon$) and then fits the model on the inliers. This is repeated multiple times to try and exhaust large number of possible configurations. In this case, a subset of 2D-3D point correspondences are used to find $[R|t]$ using E-PnP, inliers are found and new $[R|t]$ are estimated.

\par
The back projection error was less than a pixel but the $[R|t]$ we obtained was far from what we were expecting through manual tape measurements. This could mean that minimizing back projection error may not be a holistic measure in our scenario and we may have to use a better metric which relates to $[R|t]$ in a more explicit manner. In the data we collected we also introduced slight rotation. The expected rotation were calculated using trigonometry.

\begin{table}[!htbp]
\centering
\begin{tabular}{|c|c|c|c|}
\hline
\textit{rot\_data} & Expected Value & E-PnP & RanSaC E-PnP\\
 & (by tape) & 20 points & 20 points \\ \hline
X (in m) & 0.785 & 1.0346 & 1.0422 \\
Y (in m) & -0.52 & -0.557 & -0.5883 \\
Z (in m) & -0.69 & -0.393 & -0.4078 \\ \hline
Rot x (in deg) & unmeasurable & -5.82 & -6.613  \\
Rot y (in deg) & -15 & -17.54 & -17.703 \\
Rot z (in deg) & 2.6 & 3.27 & 3.126 \\ \hline
Back projection error (in pixels) & - & 4.35 & 0.5759 \\\hline
\end{tabular}
\label{tab:ransac_epnp}
\caption{Comparison of $[R|t]$ tape measurement, E-PnP, RaNSaC E-PnP. With 20 correspondences, we initialize with 15 initial points and tried 10000 different random 2D-3D point correspondence selections followed by RaNSaC and E-PnP.}
\end{table}

\vspace{7.8cm}

\begin{figure}[!htbp]
\centering
\includegraphics[width=0.65\linewidth]{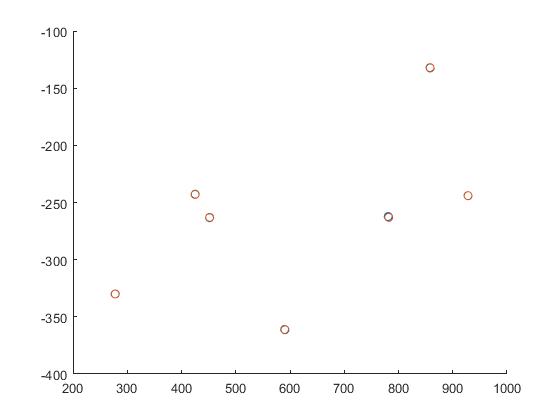}
\caption{Projection of 3D inlier points(after applying $[R|t]$ from RanSaC E-PnP) and corresponding 2D points on \textit{rot\_data}. The projected 3D and 2D points align very well but again, there is deviation from the manually measured values (refer to table \ref{tab:ransac_epnp}).}
\label{fig:min_bp_error_RANSAC_10000iters_init15pts_rot_data}
\end{figure}

\section{Using 3D-3D correspondences}
\label{3d_3d}

The 2D-3D correspondences method did not seem to work very well in our experimental setup. Error could have crept up due to not-so-accurate marking of 2D points (in pixels by looking at the image) or noisy data points to perform PnP.

The back-projection error seemed to get minimized, however, the transformation values did not seem to be in agreement with the values measured by tape.

Setups being used in real-time require extrinsic calibration to be quite accurate and produce minimal error. Fusion is one way to visualize the accuracy of the extrinsic calibration parameters. Bad calibration can result in fused data to have hallucinations in the form of duplication of objects in the fused point clouds due to bad alignment.

One such application requiring real-time fusion from multiple sensors is autonomous driving. Bad calibration can result in erroneous fused data, which can be fatal for the car as well as nearby cars, pedestrians and property.

This part of involves using augmented-reality (AR) tags and the LiDAR point cloud to find the extrinsic calibration parameters. Multiple versions of AR tags have been released by the open-source community \cite{apriltag} \cite{aruco}. The method proposed here uses the ArUco tags \cite{aruco}.

\begin{figure}[h]
\scriptsize
\centering
\setlength{\tabcolsep}{0.1em}

\begin{tabular}{ p{6cm} p{6cm} }

\includegraphics[width=0.98\linewidth]{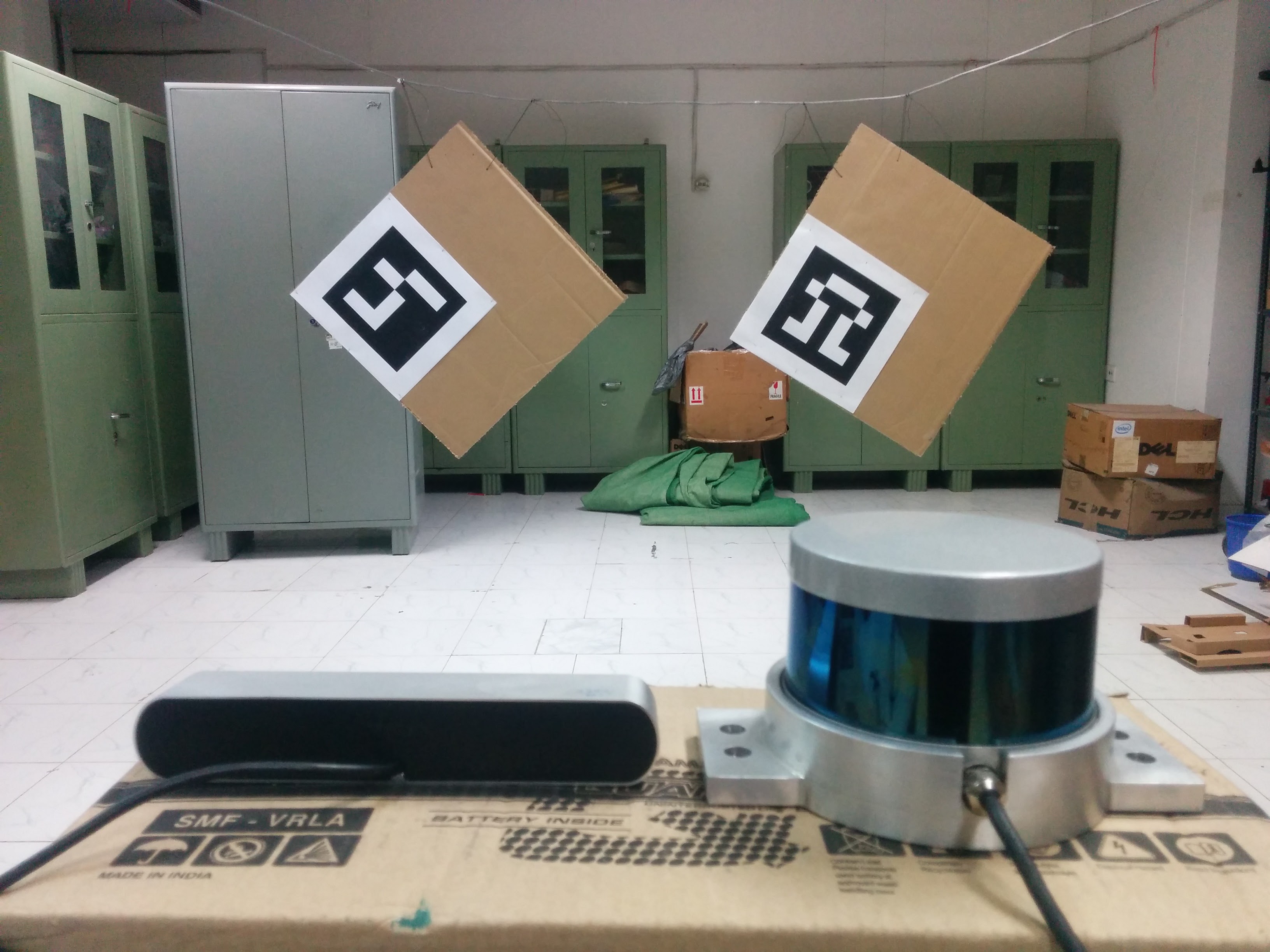} & \includegraphics[width=0.98\linewidth]{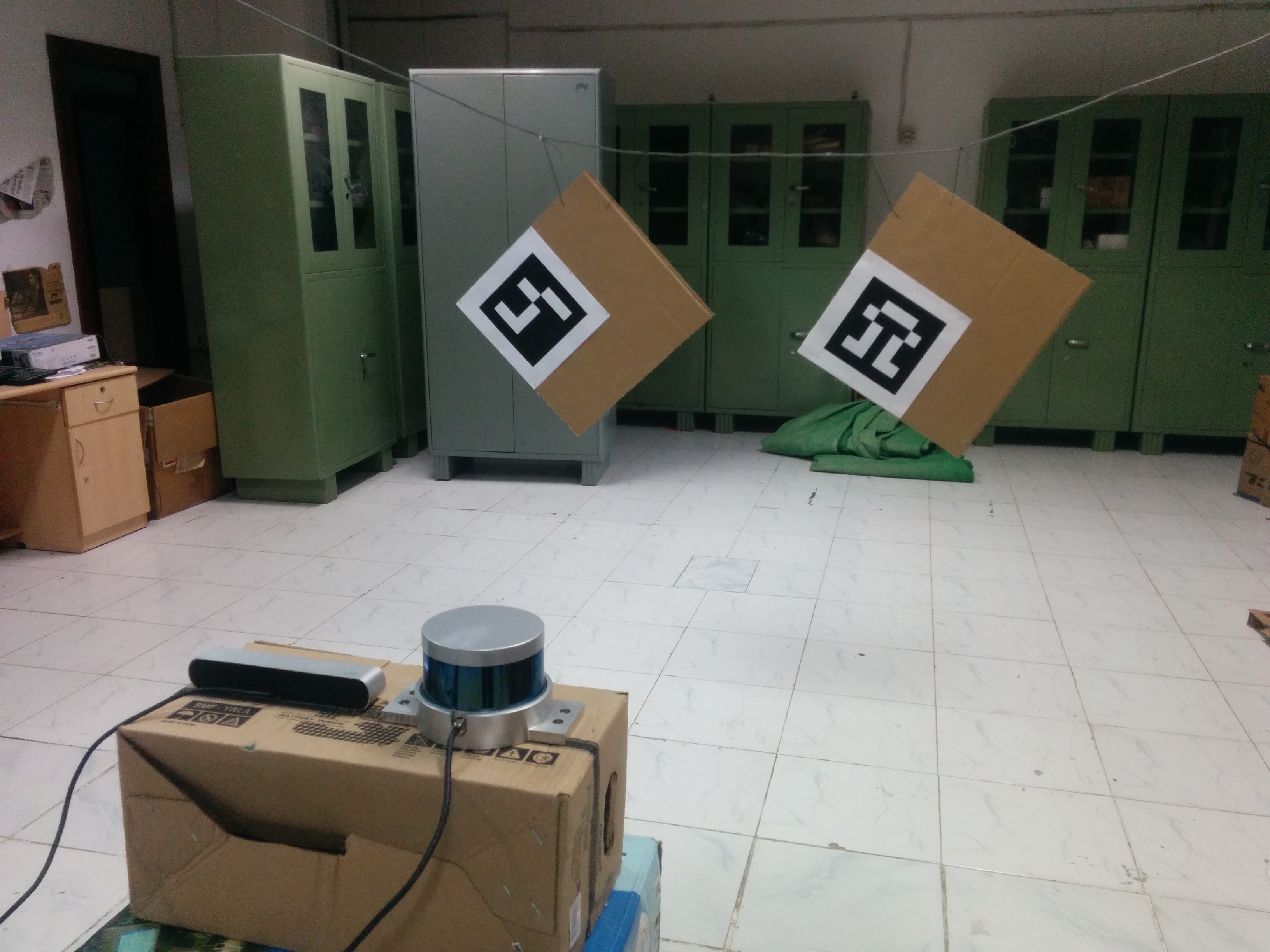}
\end{tabular}

\caption{Experimental setup with rectangular cardboard and ArUco markers using 3D-3D correspondences.}
\label{fig:3d_3d_setup}
\end{figure}

\par
To find the transformation between the camera and Velodyne, we need two sets of 3D points: one in the camera frame and another in the Velodyne frame. Once, these point correspondences are found, one can try to solve for $[R|t]$ between the two sensors.

\subsection{Experimental Setup}
Most types of calibration employ markers, dimensions, shape and specific features of which depend on the application and type of calibration being performed. Checkerboards are the most common type of markers, generally used to estimate the intrinsic calibration parameters of a camera. \cite{but_velodyne} uses special markers with circular cut-outs for calibrating a LiDAR and a camera. We have devised a pipeline which uses cost-effective markers that can be constructed easily with just a planar surface such as a cardboard and an A4 sheet of paper.

\par
The design of the marker was driven keeping in mind that it should be able to provide features/correspondences that are easy to detect, both, in the camera frame as well as the LiDAR frame.

\subsubsection{Shape and Size}
The rectangular cardboard can be of any arbitrary size. The experiments we performed used a Velodyne VLP-16\cite{velodyne} which has only 16 rings in a single scan, a handful as compared to higher density LiDARs (32 and 64 rings per scan). For a low density LiDAR, if the dimensions of the board are small and the LiDAR is kept farther than a specific distance, the number of rings hitting the board become low (2 to 3 rings resulting in only 2 to 3 points on an edge) , making it very difficult to fit lines on the edges (using RanSaC).

\par
The boards used in the experiments had length/breadth ranging between 45.0-55.0 centimeters. Keeping the LiDAR about 2.0 meters away from board with these dimensions, enough points were registered on the board edges to fit lines, calculate intersections and run the whole pipeline smoothly. It is recommended that before you run the pipeline, ensure that there are considerable number of points on the edge of the boards in the pointcloud. Any planar surface can be used: cardboards, wood or acrylic sheets. Cardboards are light-weight and can be hung easily.

\subsubsection{3D Point correspondences in the Camera Frame}
The ArUco markers are special encoded patterns that facilitate the detection and error correction of the tags themselves. More details about how they work can be found here \cite{aruco}.

\begin{figure}[h]
\vspace{-0.3cm}
\centering
\includegraphics[width=0.25\linewidth]{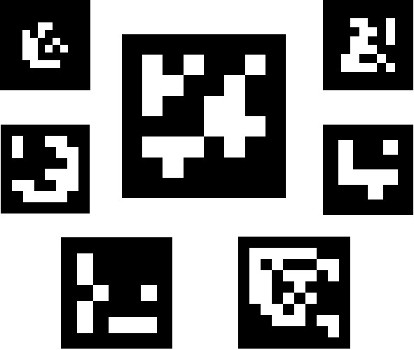}
\caption{ArUco markers.\cite{aruco}}
\label{fig:aruco_markers}
\end{figure}
\vspace{-0.5cm}

\par
The tags are stuck on a planar surface such as a rectangular cardboard. If the dimensions of the cardboard (on which the ArUco tags are stuck) and the location of the ArUco marker is known, the location of the corners (from the center of the ArUco marker) can be easily calculated.

\par
The tags provide $[R|t]$ between the camera and the center of the marker. This transform can be used to convert corner points from the marker's frame-of-reference (which is the cardboard plane with the origin being the center of the ArUcO marker) to the camera's frame-of-reference. This allows to obtain the corners as 3D points in the camera frame. We used ZED stereo camera \cite{zed}.

\subsubsection{3D Point correspondences in the LiDAR Frame}
Points in the LiDAR can be found by detecting edges of the cardboard, which in turn can be solved for corners in a similar fashion described in Section \ref{2d_3d}.

\begin{figure}[h]
\centering
\includegraphics[width=0.7\linewidth]{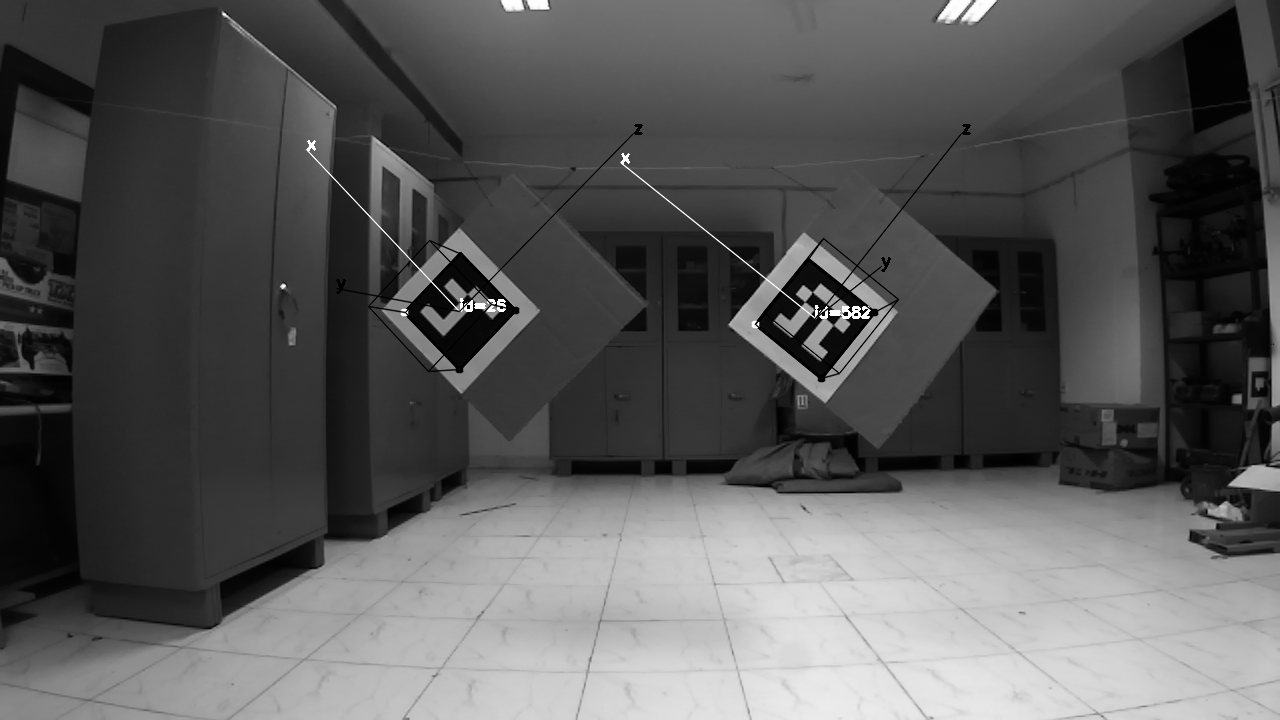}
\caption{Rotation and translation between the board's frame-of-reference and the camera's frame-of-reference are found using the ArUco markers. This transformation is used to transform the 3D corner points from the board frame to the camera frame.}
\label{fig:aruco_setup}

\end{figure}

\par
The values of transformations obtained using ArUco markers, especially the translation was quite accurate and close to values measured by tape between the camera and the center of each marker. Once the two sets of point correspondences are obtained, $[R|t]$ between their co-ordinate frames can be estimated using the Iterative Closest Point (ICP) algorithm \cite{icp}. The ICP tries to minimize the error in 3D and is given by equation \ref{eq:icp_error}.

\begin{equation} \label{eq:icp_error}
\argmin_{R \in SO(3), t \in \mathbb{R}^{3}} ||(RP+t)-Q||^{2}
\end{equation}

\par
The general ICP algorithm considers the closest points in both the point clouds as correspondences (there are other variants of choosing the closest points), following which, it finds the $[R|t]$ which best align the two point clouds by minimizing the euclidean distance between corresponding points.

\par
Finding the right correspondences can be tricky and may lead to an undesired solution. Since, in our proposed method the point correspondences are known, the corners of the marker in this case, a closed form solution exists. The Kabsch algorithm\cite{kabsch} \cite{molecular_distance_measures} finds the rotation between two point clouds and the translation can be found once the co-ordinate frames are aligned.

\par
Using the same arguments as in \cite{molecular_distance_measures}. First, we assume that the rotation is known and solve for the translation between the two point clouds, $P$ and $Q$.

\begin{equation} \label{eq:icp_error}
F(t) = \sum_{i=1}^{n} ||(RP_{i}+t)-Q_{i}||^{2}
\end{equation}

\begin{equation} \label{eq:icp_error2}
\frac{\partial F(t)}{\partial t} = 2 \sum_{i=1}^{n} (RP_{i}+t)-Q_{i} = 0
\end{equation}

\begin{equation} \label{eq:icp_error3}
\frac{\partial F(t)}{\partial t} = 2 R \sum_{i=1}^{n} P_{i} + 2 t \sum_{i=1}^{n} 1 - 2 \sum_{i=1}^{n} Q_{i}
\end{equation}

\begin{equation} \label{eq:icp_error4}
t = \frac{1}{n} \sum_{i=1}^{n} Q_{i} - R \frac{1}{n} \sum_{i=1}^{n} P_{i}
\end{equation}

\begin{equation} \label{eq:icp_error5}
t = \bar{Q} - R \bar{P}
\end{equation}

Substituting the result of equation \ref{eq:icp_error5} in objective function \ref{eq:icp_error}.

\begin{equation} \label{eq:icp_error6}
R = \argmin_{R \in SO(3)} ||(R(P_{i}-\bar{P})-(Q_{i}-\bar{Q})||^{2}
\end{equation}

let,
\begin{equation} \label{eq:icp_error7}
X = (P_{i}-\bar{P}) \textrm{ , }
X' = RX \textrm{ and }
Y = (Q_{i}-\bar{Q})
\end{equation}

then, the objective becomes,

\begin{equation} \label{eq:icp_error8}
\sum_{i=1}^{n} ||X'_{i} - Y_{i}||^{2} = Tr((X'-Y)^{T}(X'-Y))
\end{equation}

using properties of the trace of a matrix, the above equation can be simplified as,

\begin{equation} \label{eq:icp_error9}
Tr((X'-Y)^{T}(X'-Y)) = Tr(X'^{T}X') + Tr(Y^{T}Y) - 2Tr(Y^{T}X))
\end{equation}

since, the $R$ is an orthonormal matrix, it preserves lengths i.e. $|X'_{i}|^{2} = |X_{i}|^{2} $,

\begin{equation} \label{eq:icp_error10}
Tr((X'-Y)^{T}(X'-Y)) = \sum_{i=1}^{n} ( |X_{i}|^{2} + |Y_{i}|^{2} ) - 2Tr(Y^{T}X')
\end{equation}

re-writing the objective function by eliminating terms that do not involve $R$,

\begin{equation} \label{eq:icp_error11}
R = \argmax_{R \in SO(3)} \hspace{2mm} Tr(Y^{T}X')
\end{equation}

substituting the value of $X'$ and using property of the trace,
\begin{equation} \label{eq:icp_error12}
Tr(Y^{T}X') = Tr(Y^{T}RX) = Tr(XY^{T}R)
\end{equation}

using SVD on $XY^{T} = UDV^{T}$,
\begin{equation} \label{eq:icp_error13}
Tr(XY^{T}R) = Tr(UDV^{T}R) = Tr(DV^{T}RU) = \sum_{i=1}^{3} d_{i}v_{i}^{T}Ru_{i}
\end{equation}

let, $M = V^{T}RU$, then,
\begin{equation} \label{eq:icp_error14}
Tr(Y^{T}X') = \sum_{i=1}^{3} d_{i}M_{ii} \leq \sum_{i=1}^{3} d_{i}
\end{equation}

$M$ is a product of orthonormal matrices and is an orthonormal matrix as well with $det(M)=+/-1$. The length of each column vector in $M$ is equal to one and each component of a vector is less than or equal to one. Now, to maximize the above equation, let each $M_{ii}=1$, forcing the remaining components of the vector to zero to satisfy the unit vector constraint. Thus, $M=I$, an identity matrix.

\begin{equation} \label{eq:icp_error15}
M=I \implies V^{T}RU=I \implies R=VU^{T}
\end{equation}

to ensure, that $R$ is a proper rotation matrix, i.e. $R \in SO(3)$, we need to make sure that $det(R)=+1$. If the $R$ obtained from equation \ref{eq:icp_error15} has $det(R)=-1$ we need to find $R$ such that $Tr(Y^{T}X')$ takes the second largest value possible.

\begin{equation} \label{eq:icp_error16}
Tr(Y^{T}X') = d_{1}M_{11} + d_{2}M_{22} + d_{3}M_{33} \textrm{ where } d{1} \geq d{2} \geq d{3} \textrm{ and } |M_{ii}| \leq 1
\end{equation}

the second largest value of the term in equation \ref{eq:icp_error16} occurs when $M_{11} = M_{22} = +1$ and $M_{33} = -1$. Taking into account the above,

\begin{equation} \label{eq:icp_error17}
R = UCV^{T}
\end{equation}

where $C$ is a correction matrix,

\begin{equation} \label{eq:icp_error18}
C =
\begin{bmatrix}
    1 & 0 & 0 \\
    0 & 1 & 0 \\
    0 & 0 & sign(det(UV^{T})) \cdot 1 \\
\end{bmatrix}
\end{equation}

\subsection{Incorporating multiple scans}
\label{multi_scans}
In our initial experiments, we observed that even in a closed room where the boards are as stationary as can be, the pointcloud visualized in Rviz shows that the points (from the LiDAR), on the contrary, are not stationary and there is a small amount of position shift between two instants.

\par
To reduce any noise that might creep up we further propose to collect multiple samples of rotations and translations (using the method discussed above). Rotations and translation estimated over multiple runs can be used to obtain a more accurate and less noisy rigid-body transformation that transforms points from the LiDAR frame to the camera frame. Multiple sensor data, is collected over $N$ iterations, keeping the positions of the LiDAR and camera fixed.

\par
From each one of the $N$ runs we estimate the rotation and translation. We can average the $N$ observed translation vectors,

\begin{equation} \label{eq:tvec_avg}
\bar{t} = \frac{1}{N} \sum_{i=1}^{N} tvec_{i}
\end{equation}

where $tvec_{i} \in \mathbb{R}^{3}$ and $\bar{t}$ is the average translation between the two sensors.

\par
\begin{figure}[!htbp]
\vspace{0.0cm}
\scriptsize
\centering
\setlength{\tabcolsep}{0.1em}

\begin{tabular}{ p{4cm} p{4cm} p{4cm} }

\includegraphics[width=0.98\linewidth]{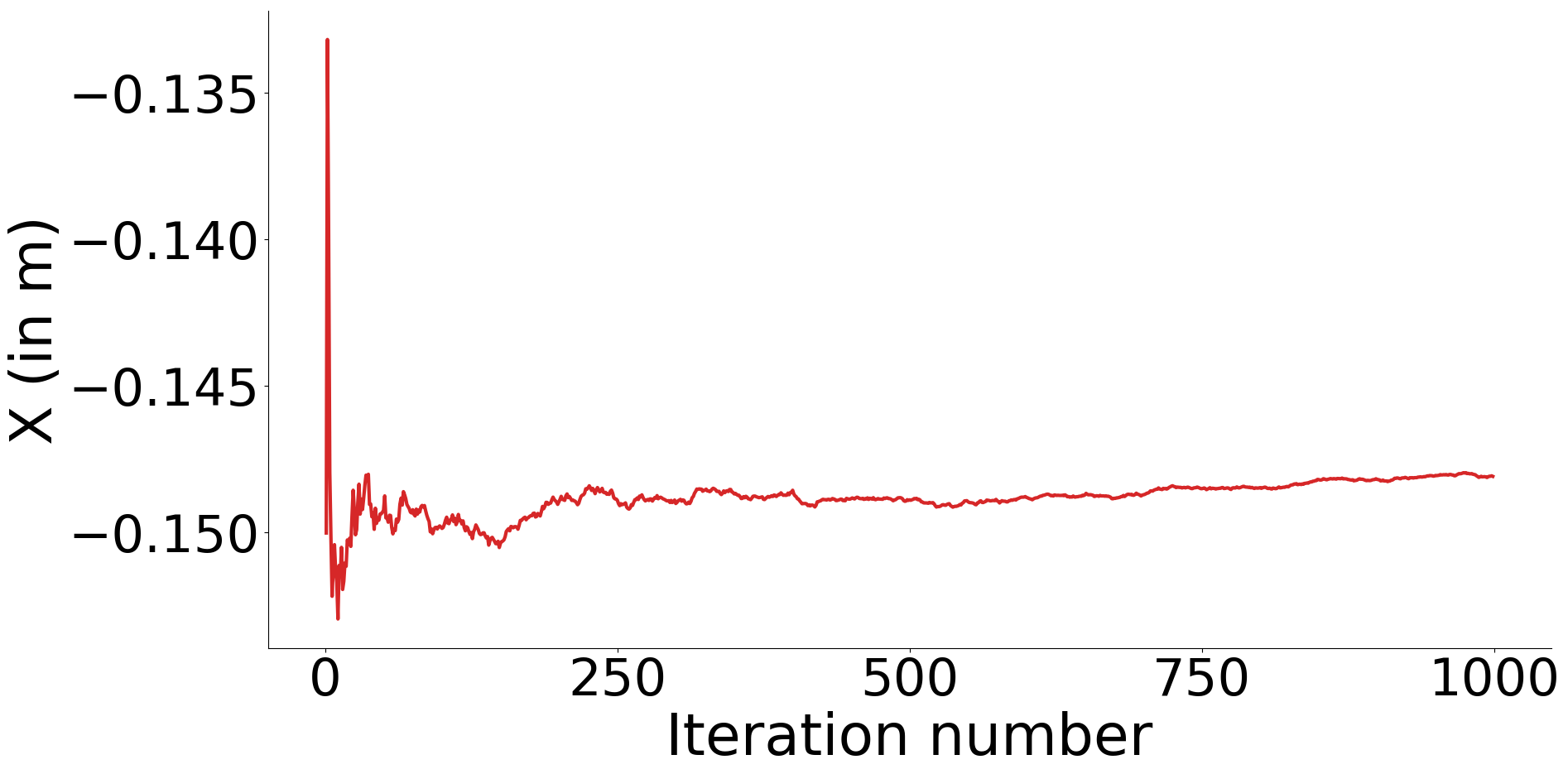} &
\includegraphics[width=0.98\linewidth]{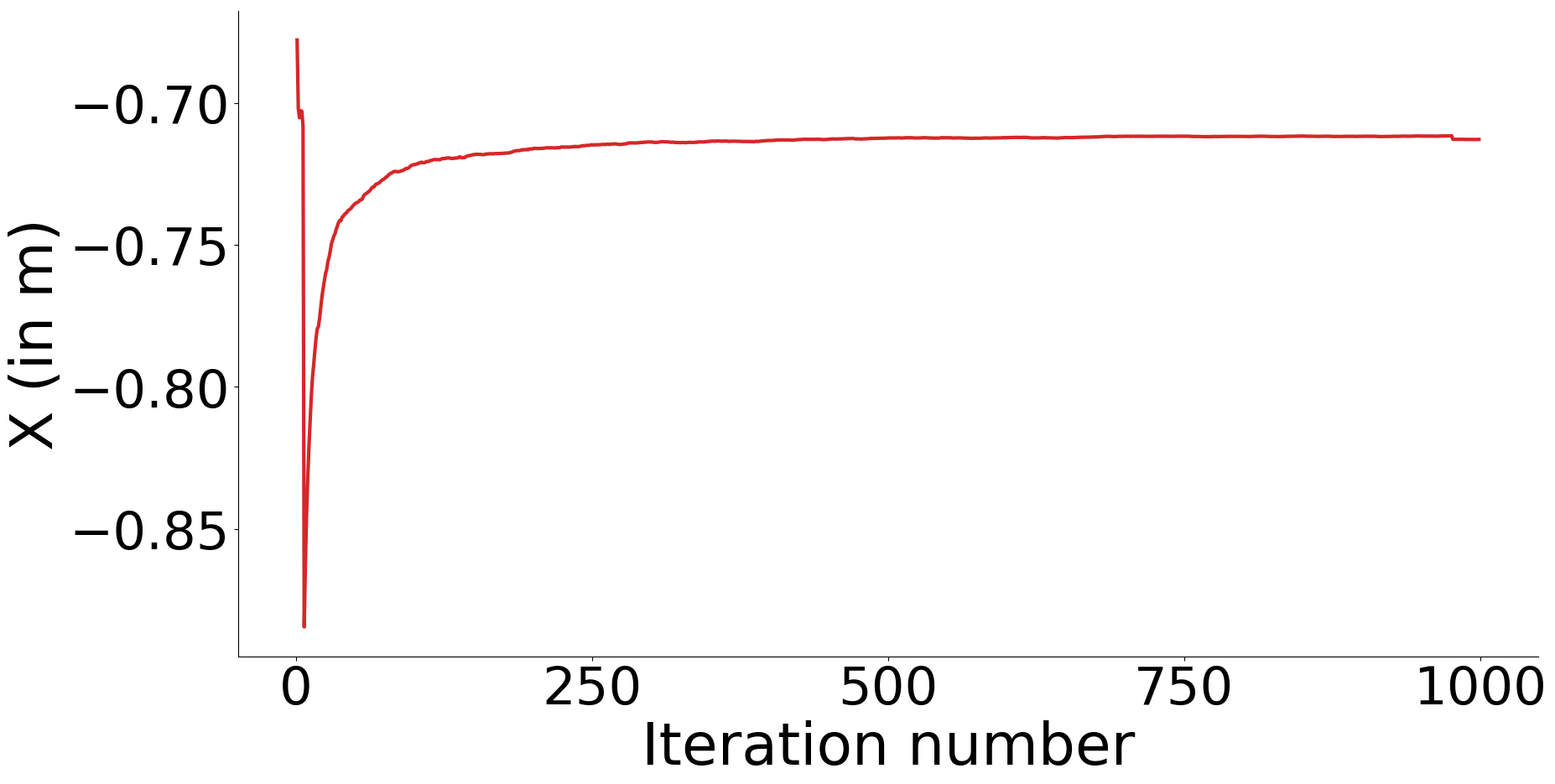} & \includegraphics[width=0.98\linewidth]{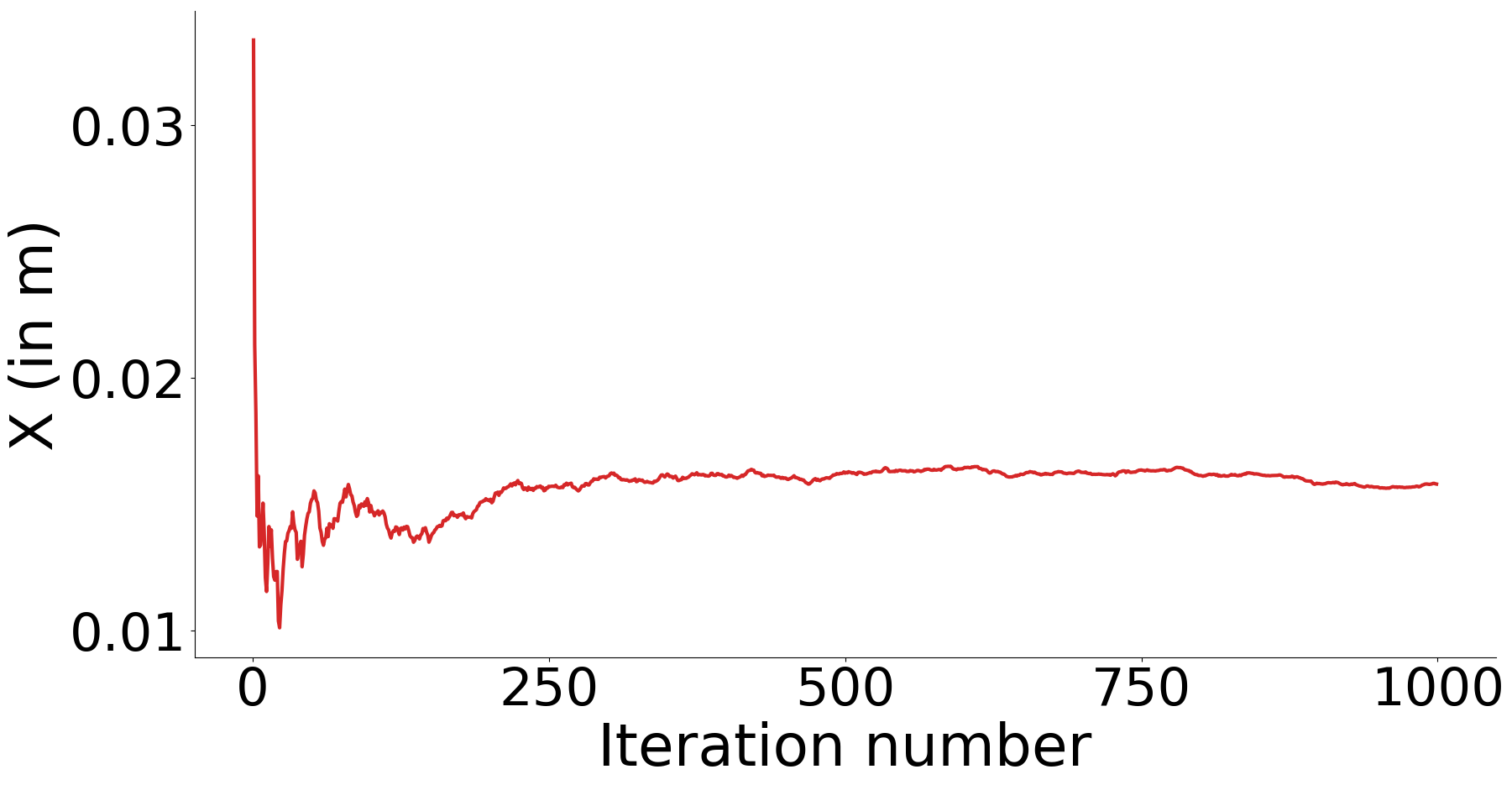} \\

\includegraphics[width=0.98\linewidth]{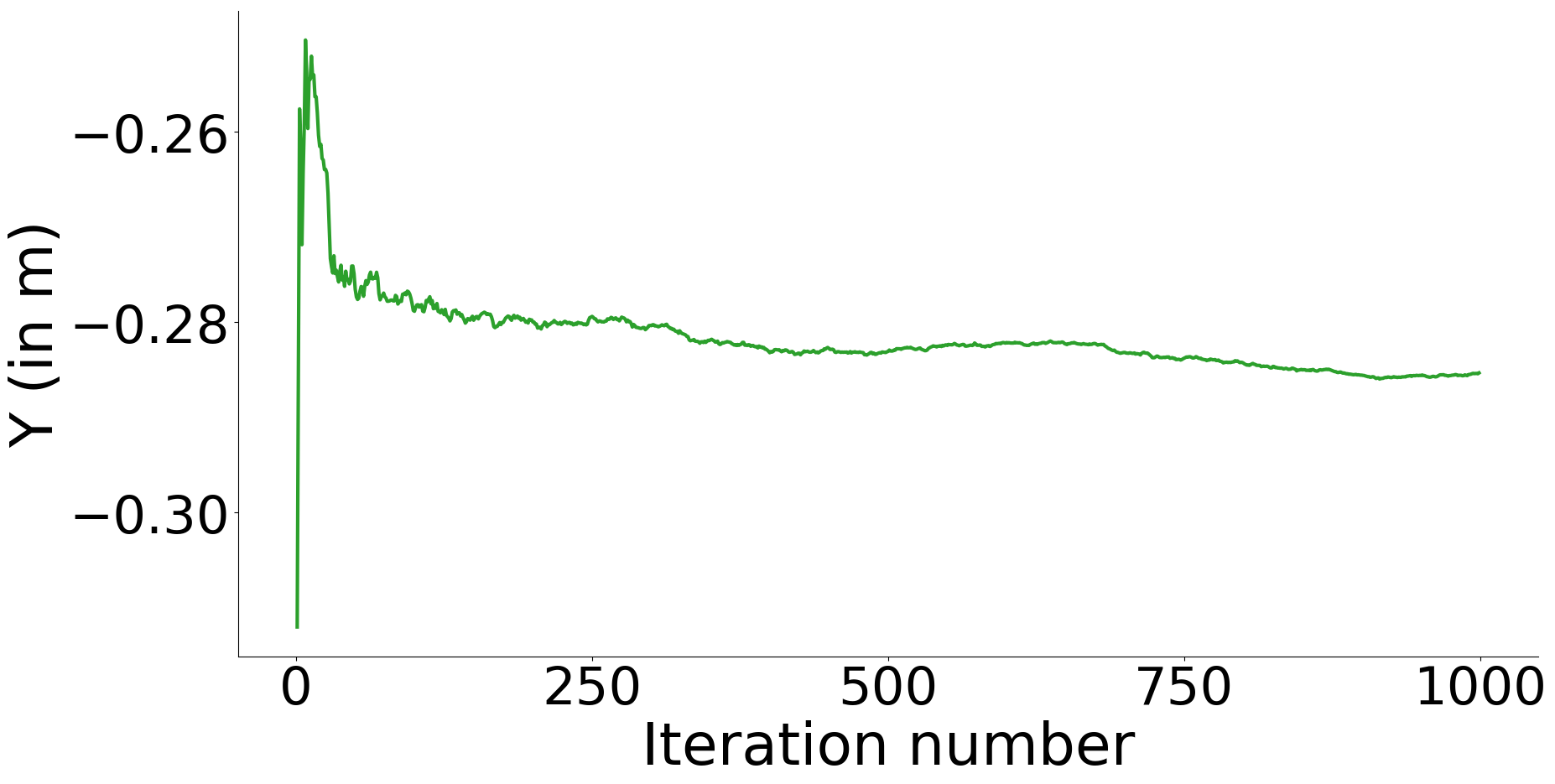} &
\includegraphics[width=0.98\linewidth]{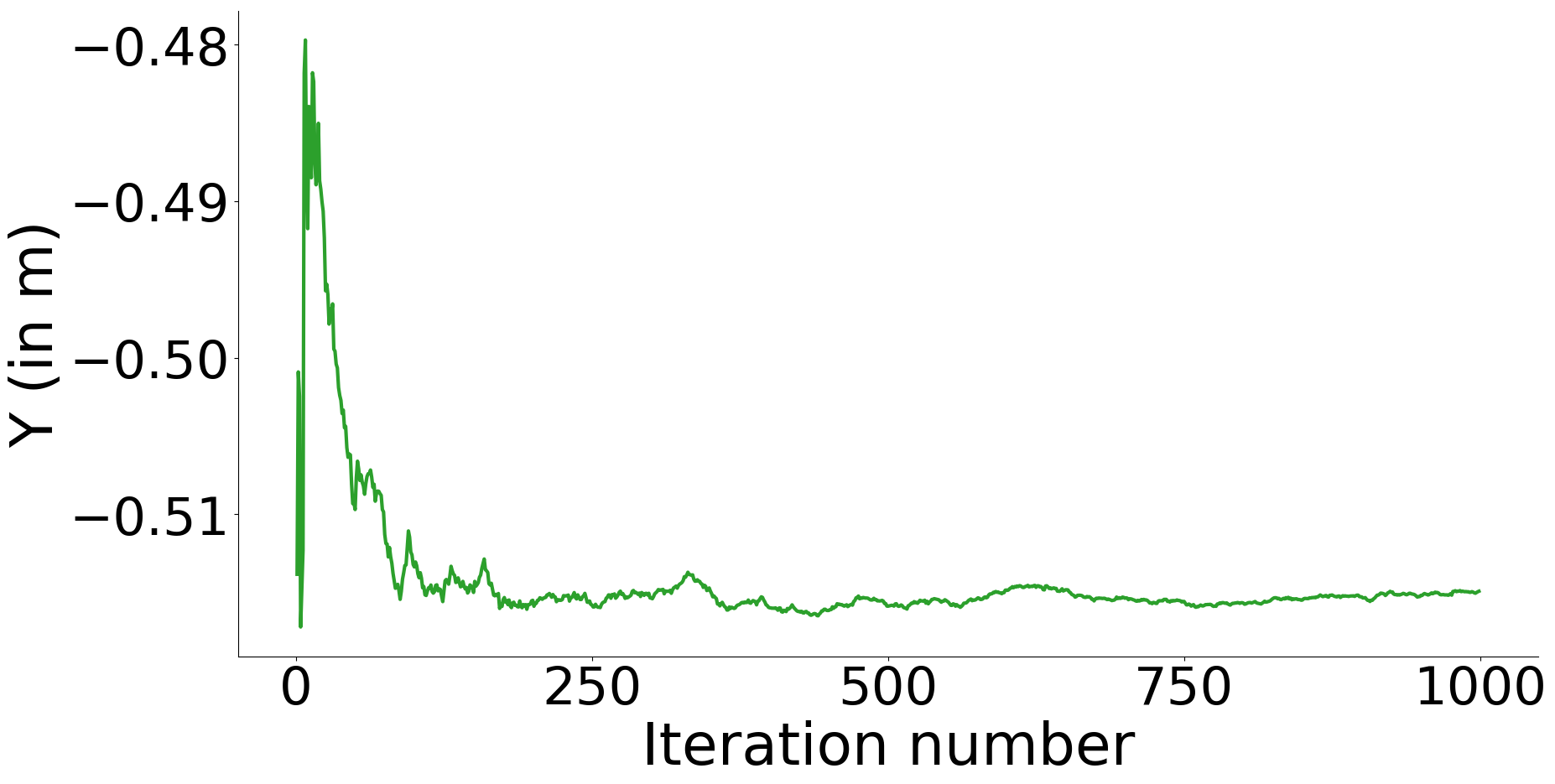} & \includegraphics[width=0.98\linewidth]{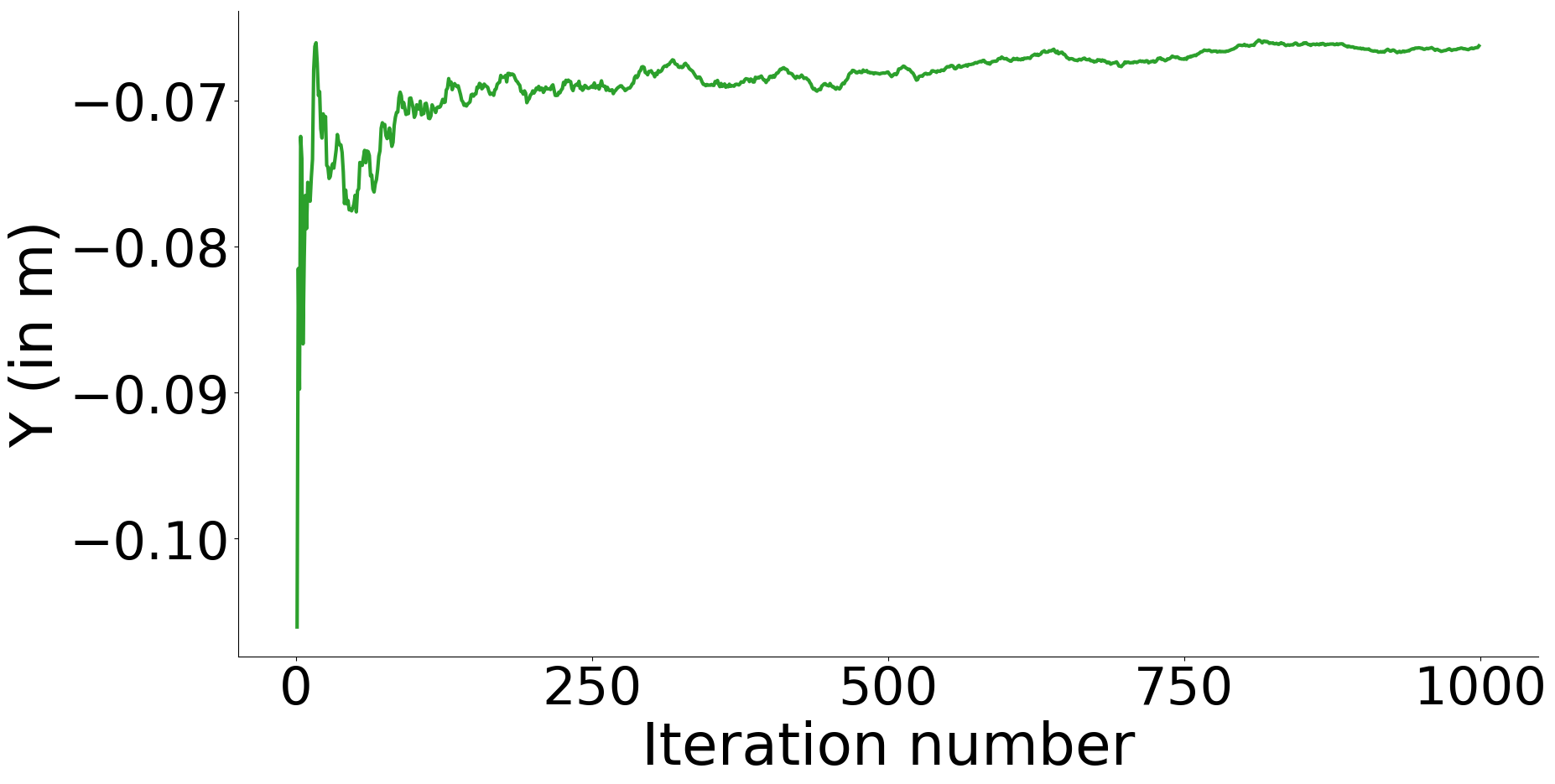} \\

\includegraphics[width=0.98\linewidth]{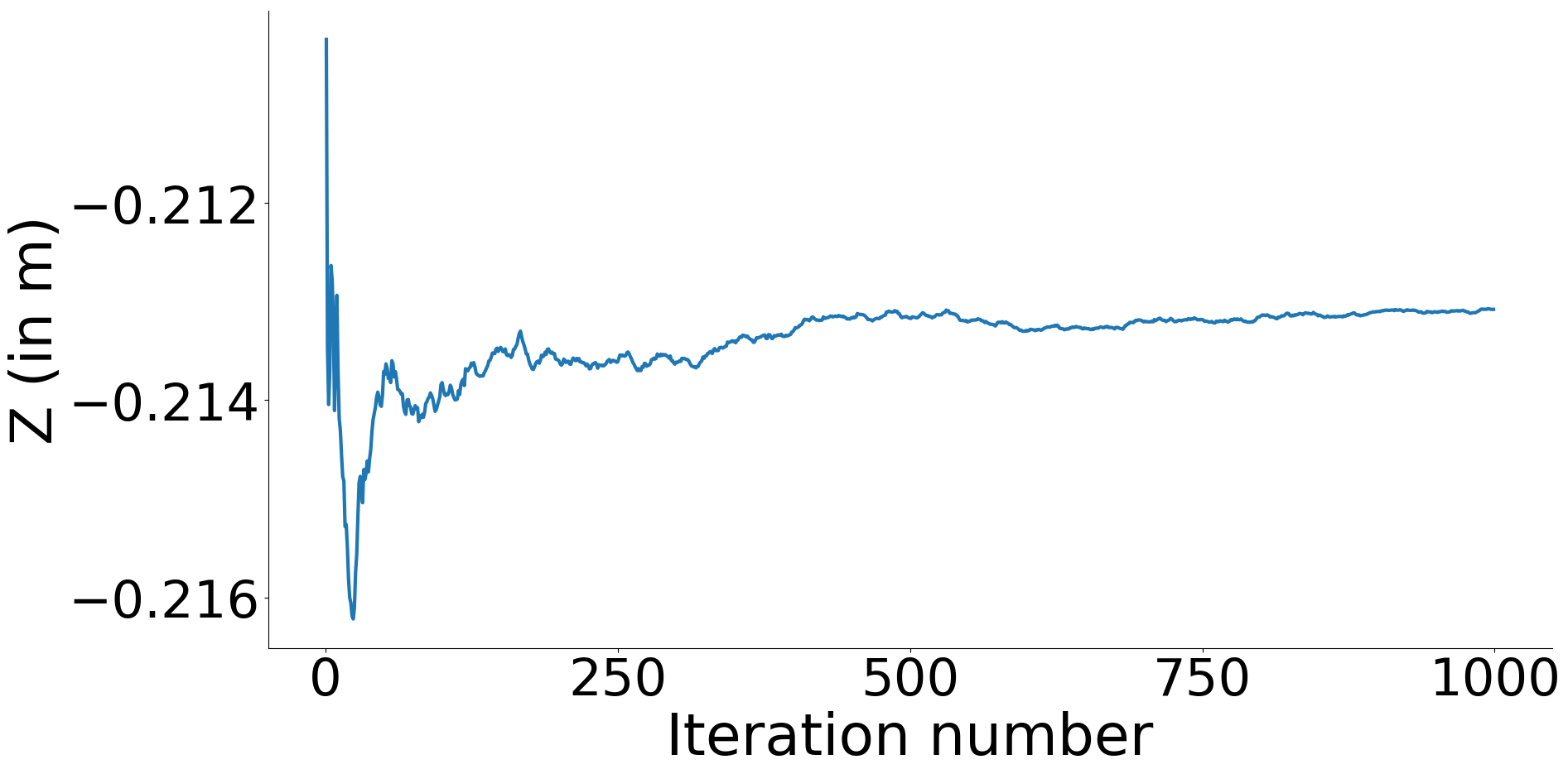} &
\includegraphics[width=0.98\linewidth]{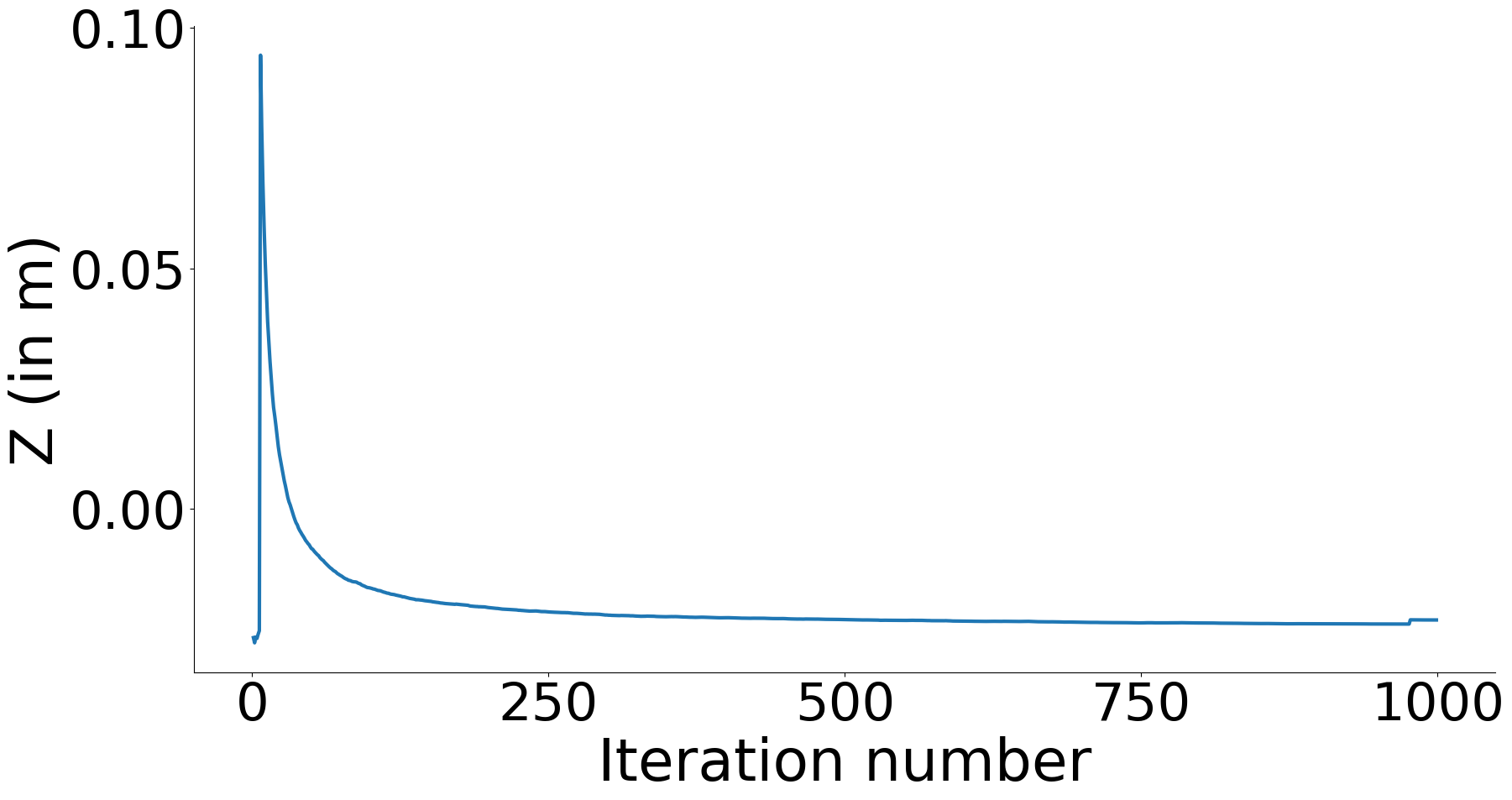} & \includegraphics[width=0.98\linewidth]{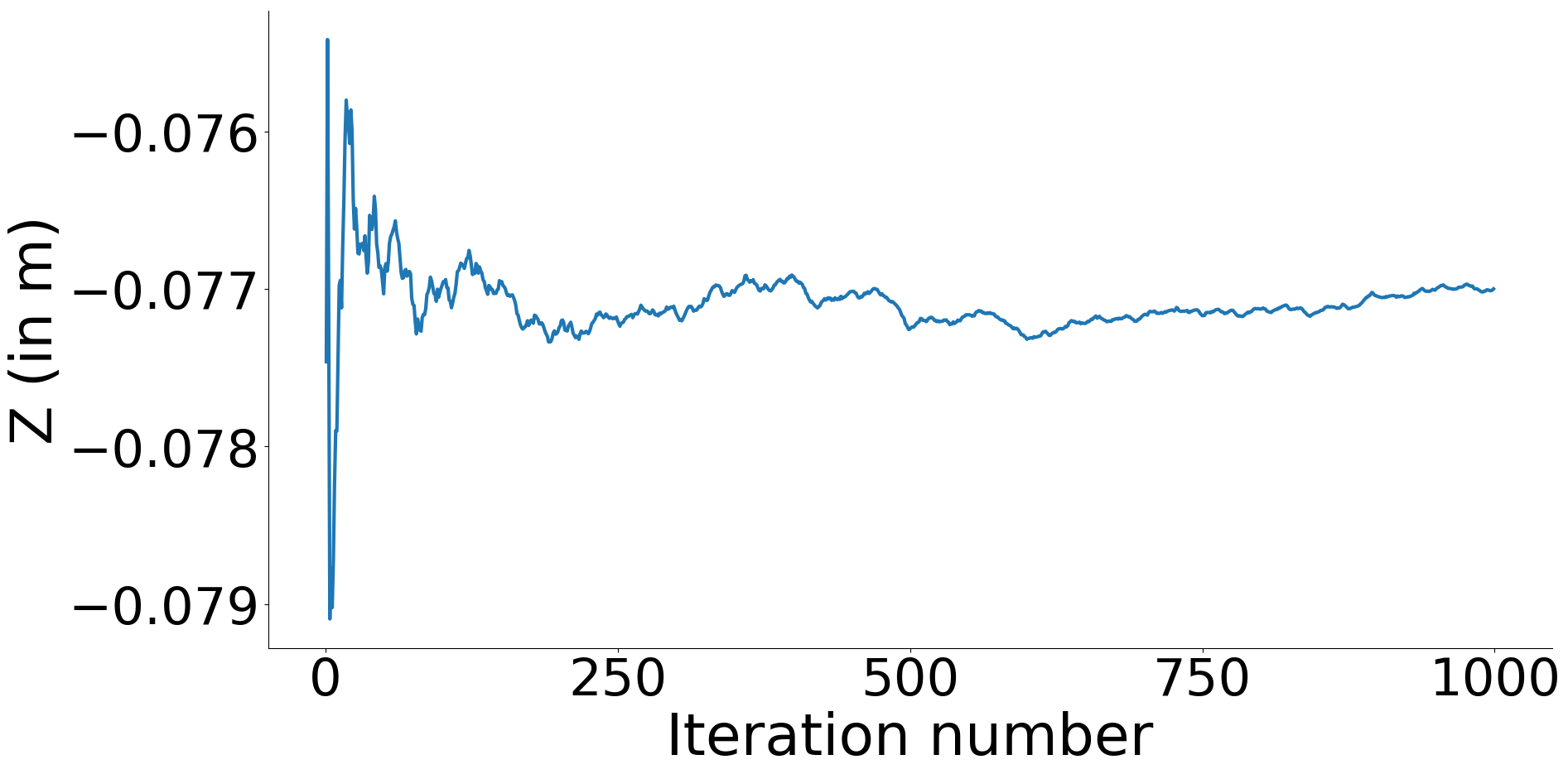} \\

\centering
(a) Sensor Configuration 1 &
\centering
(b) Sensor Configuration 2  \\[6pt] &
\centering
(c) Sensor Configuration 3  \\[6pt]

\end{tabular}

\vspace{-0.4cm}
\caption{At the $i$-th iteration, the plots show the translation values after averaging estimates from the previous $i$-iterations. The plots are show translation in $X, Y, Z$ when the LiDAR and camera were kept in three random configurations(positions).}
\label{fig:multiple_scans}
\end{figure}

Taking average of rotation matrices is not very straight-forward, so we transform them to quaternions, compute the average quaternion in $\mathbb{R}^{4}$ and then convert it back to rotation matrix.

\begin{equation} \label{eq:rvec_avg}
r = \frac{1}{N} \sum_{i=1}^{N} rvec_{i}
\end{equation}

\begin{equation} \label{eq:rvec_avg}
\bar{r} = \frac{r}{||r||}
\end{equation}

where $rvec_{i}$ is a quaternion representation of rotation matrix obtained in the $i$th run and $\bar{r}$ is the average rotation between the two sensors represented by a unit-quaternion. The results of averaging in three separate configurations for LiDAR and camera positions can be seen in figure \ref{fig:multiple_scans}.

Averaging multiple estimates while keeping the position of the relative positions of the LiDAR and camera fixed helps to reduce noisy data due to imperfect marker edges and errors that might be introduced due to LiDAR points being slightly inaccurate. Also, if there is a modest amount of motion of the cardboards, we effectively observe many data points around the actual translation and rotation, averaging which shall give a better estimate of the rigid-body transformation between the LiDAR and camera.

\section{Fusing point clouds}
\label{fusing_pcl}
Our main objective was to accurately calibrate multiple cameras that may not have an overlapping field-of-view. If a set of transformations can be estimated that transform all sensor data to a single frame of reference, data, such as point clouds from stereo cameras can be fused. A setup with multiple stereo cameras facing in different directions, one can obtain a point cloud that provides a 360-degree field-of-view by fusing individual point clouds from each stereo camera.

\begin{figure}[!htbp]
\vspace{-0.6cm}
\scriptsize
\centering
\setlength{\tabcolsep}{0.1em}

\begin{tabular}{ p{6cm} p{6cm} }

\includegraphics[width=0.98\linewidth]{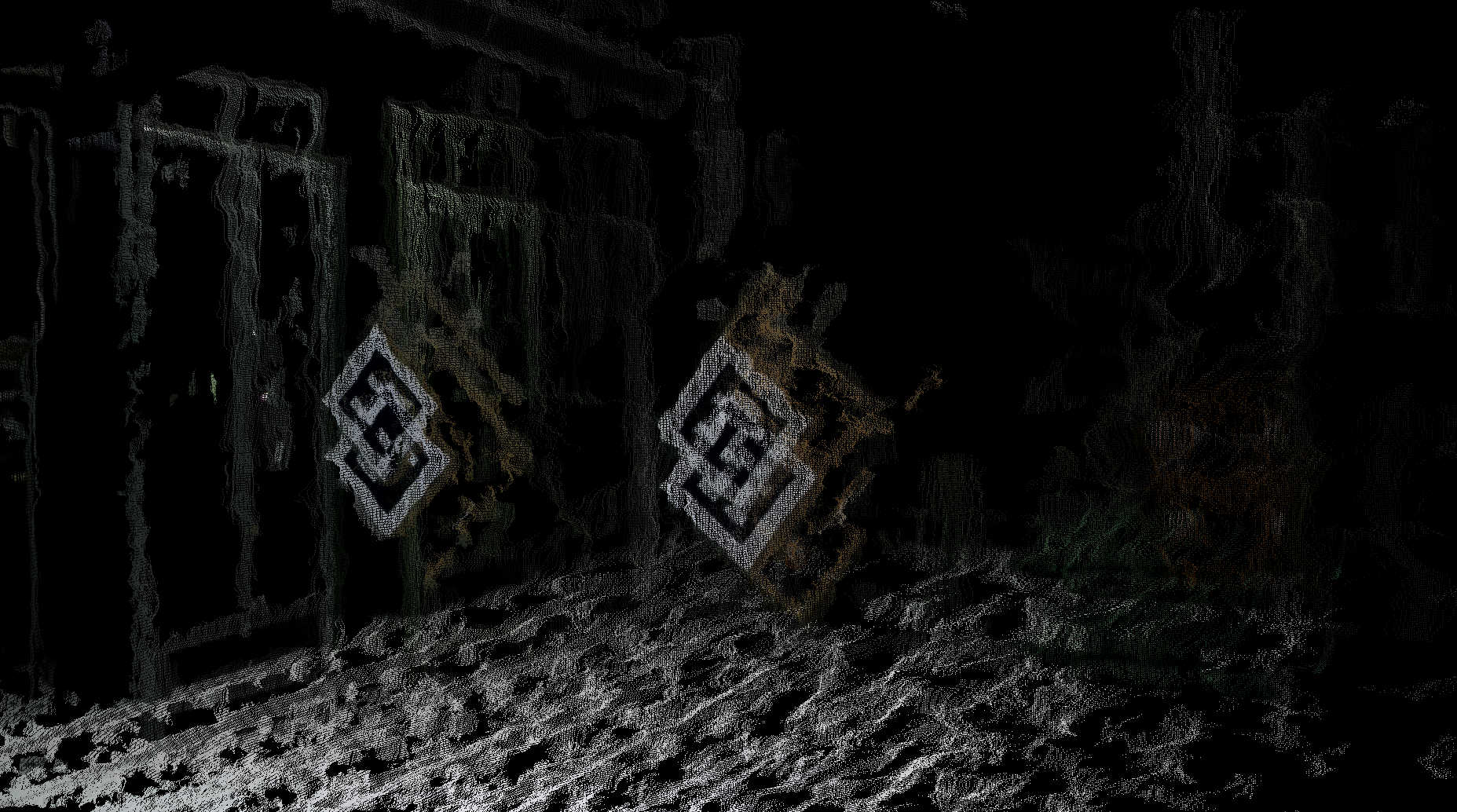} & \includegraphics[width=0.98\linewidth]{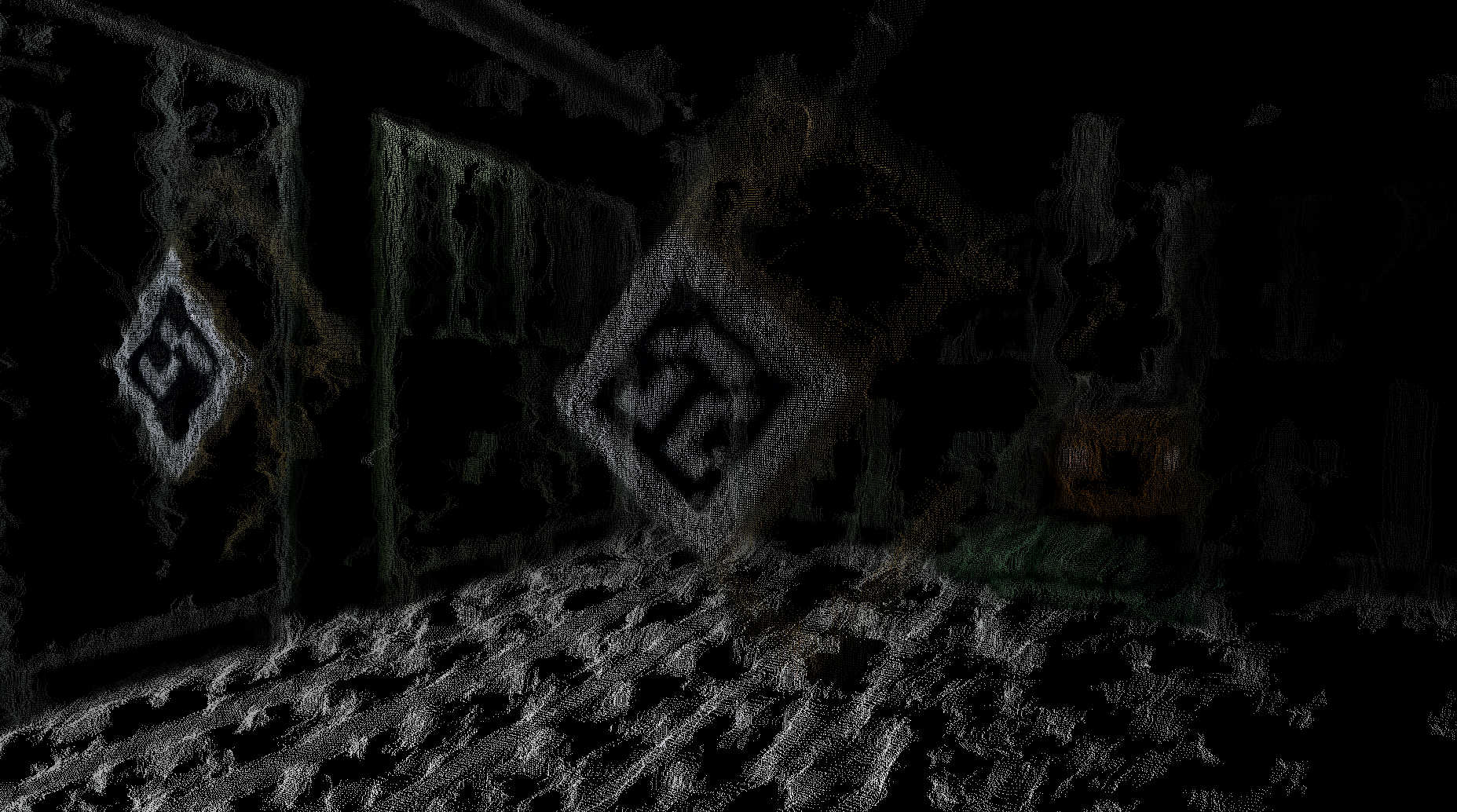} \\

\includegraphics[width=0.98\linewidth]{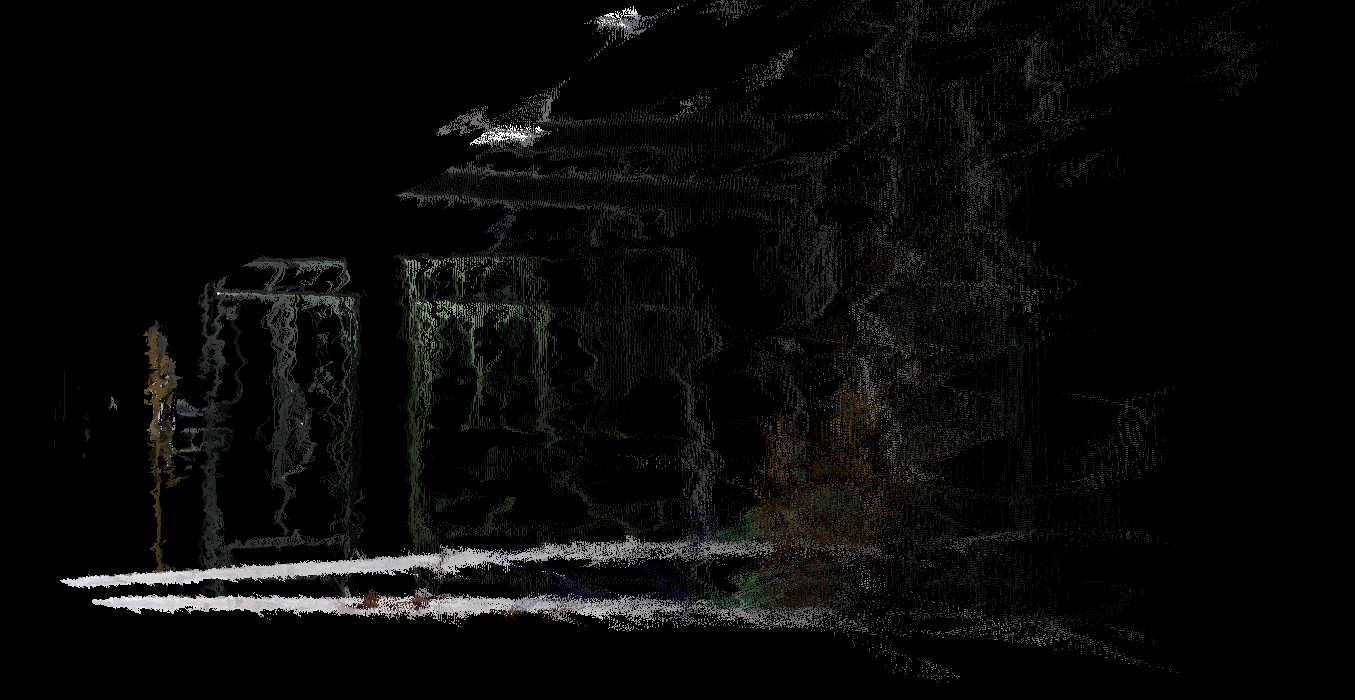} & \includegraphics[width=0.98\linewidth]{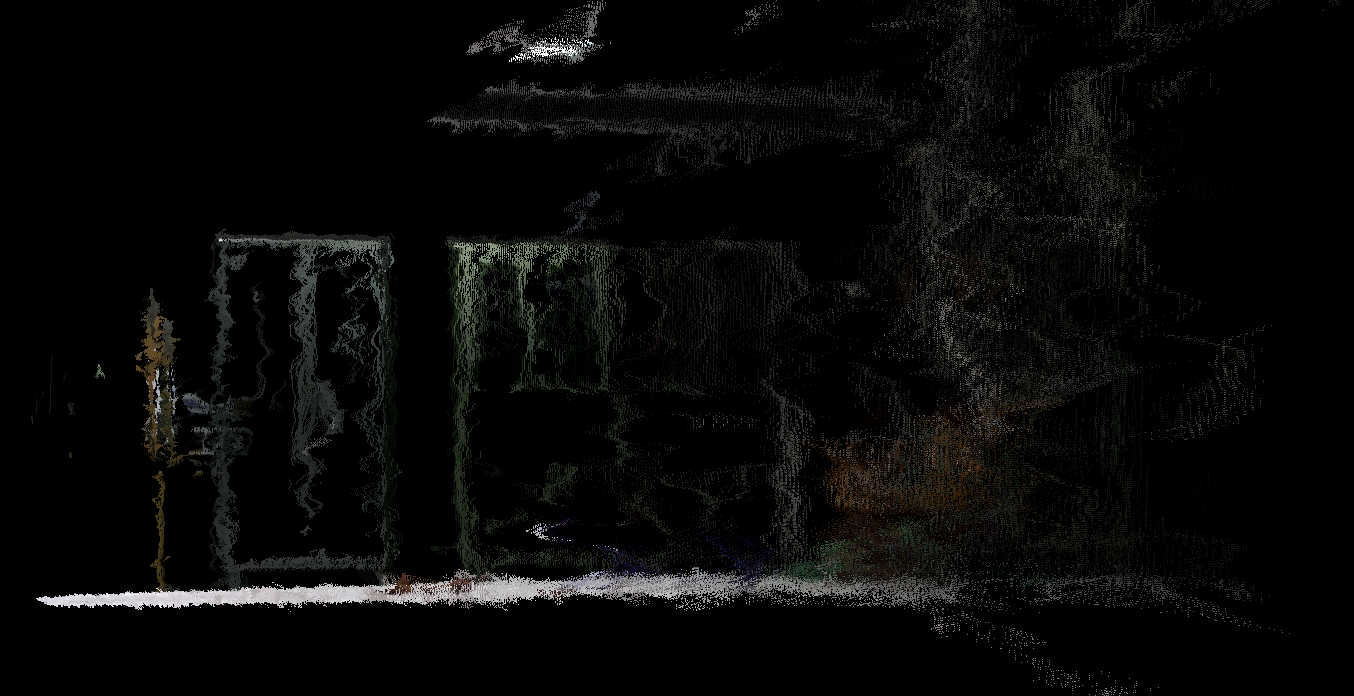} \\

\includegraphics[width=0.98\linewidth]{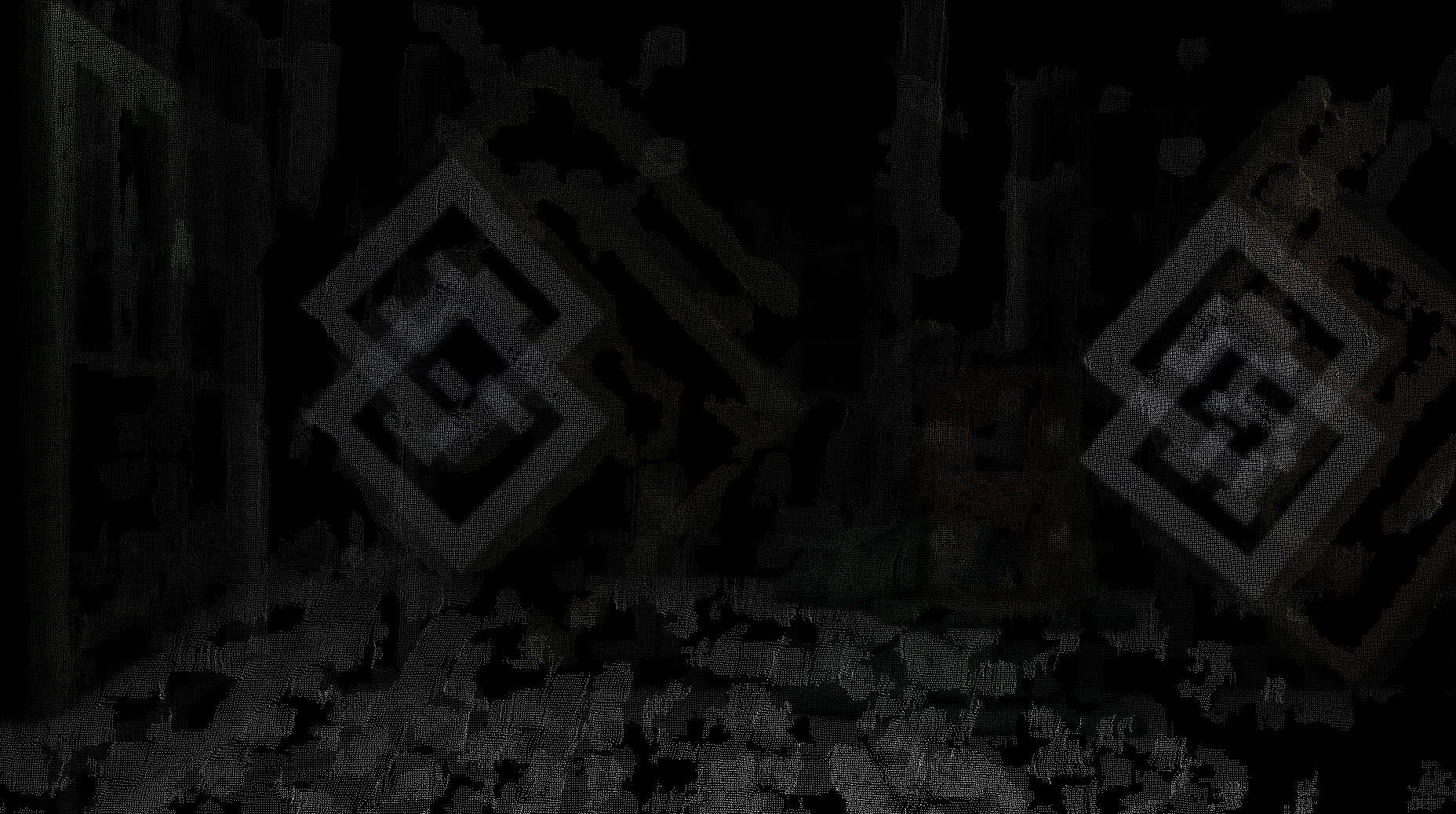} & \includegraphics[width=0.98\linewidth]{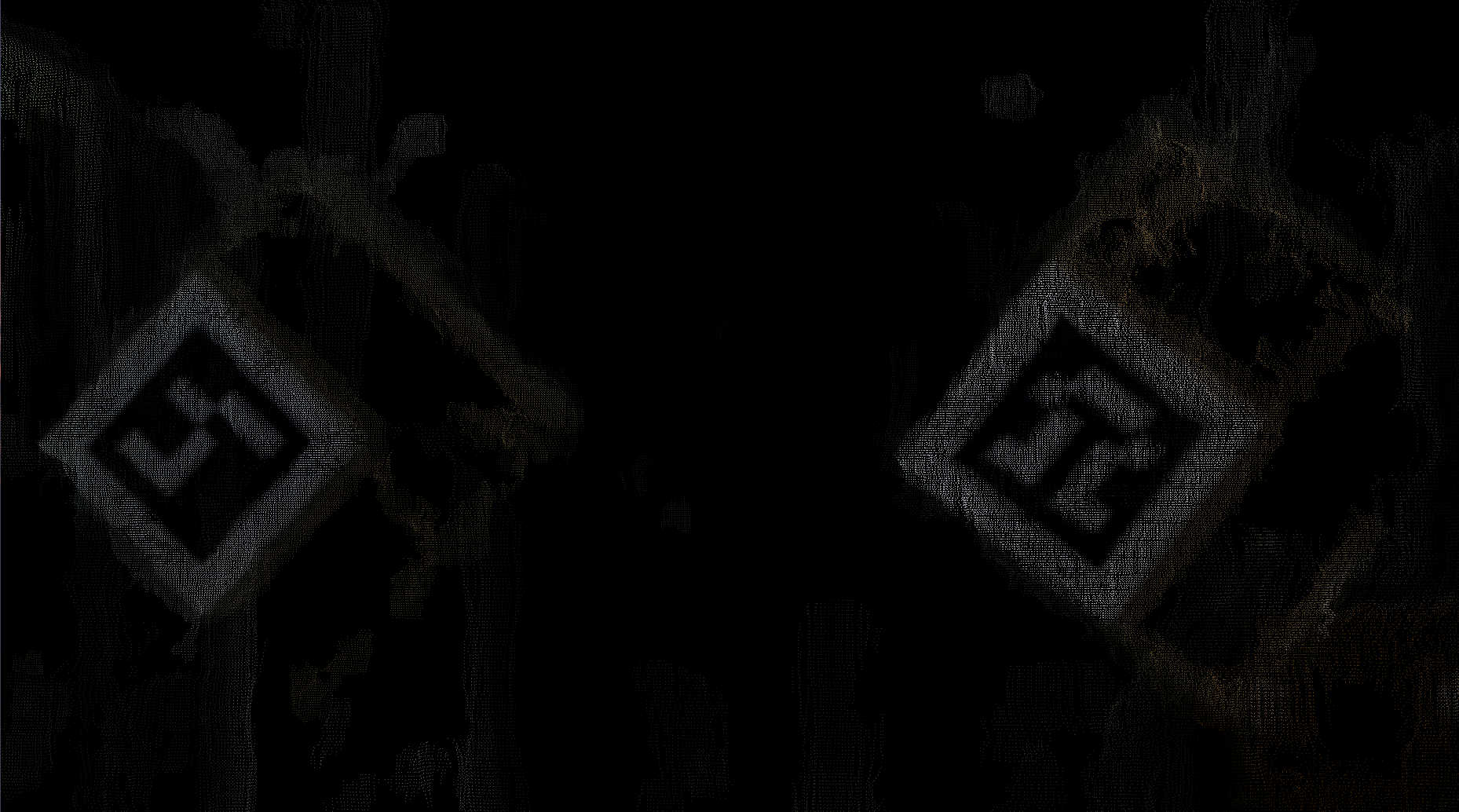} \\

\includegraphics[width=0.98\linewidth]{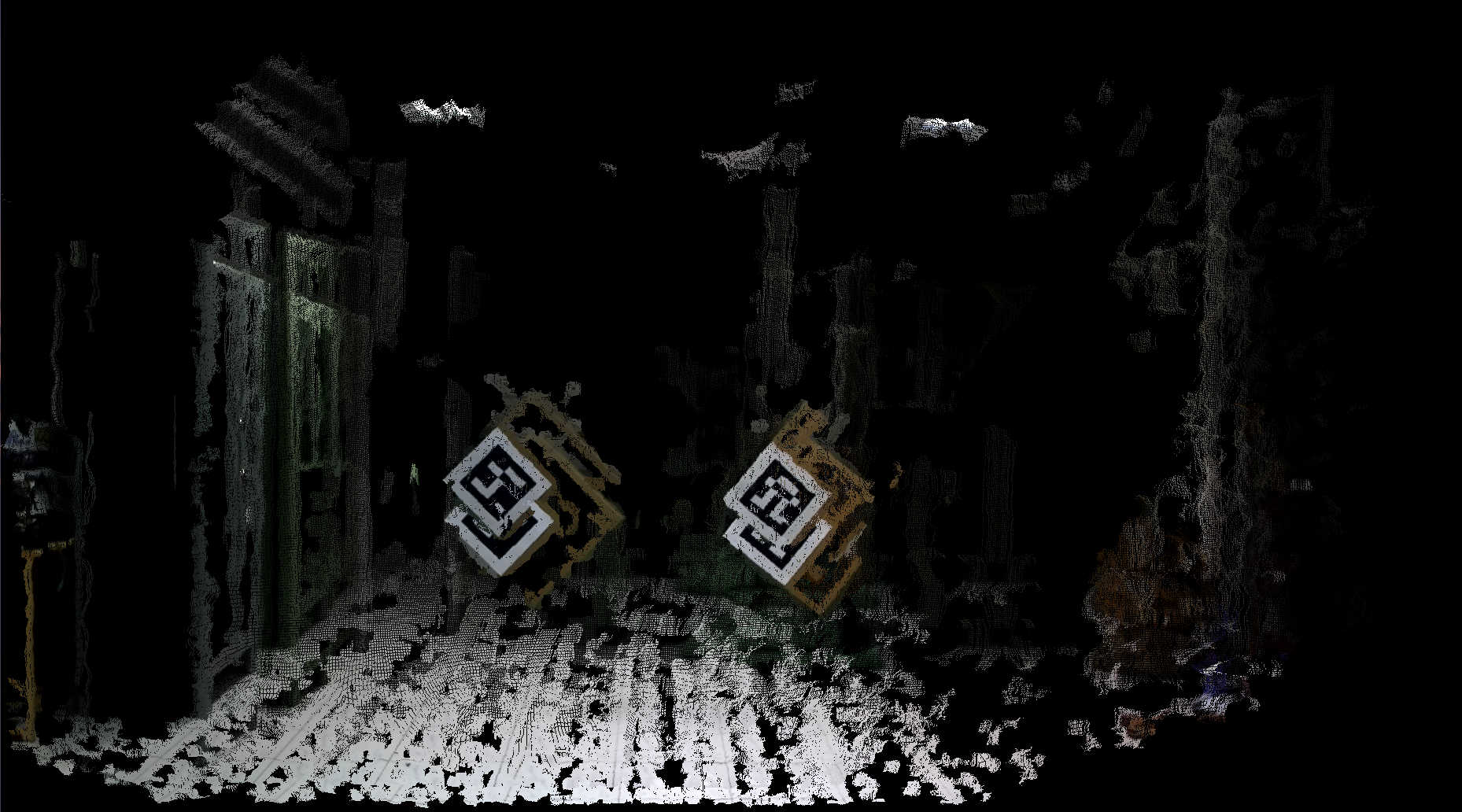} & \includegraphics[width=0.98\linewidth]{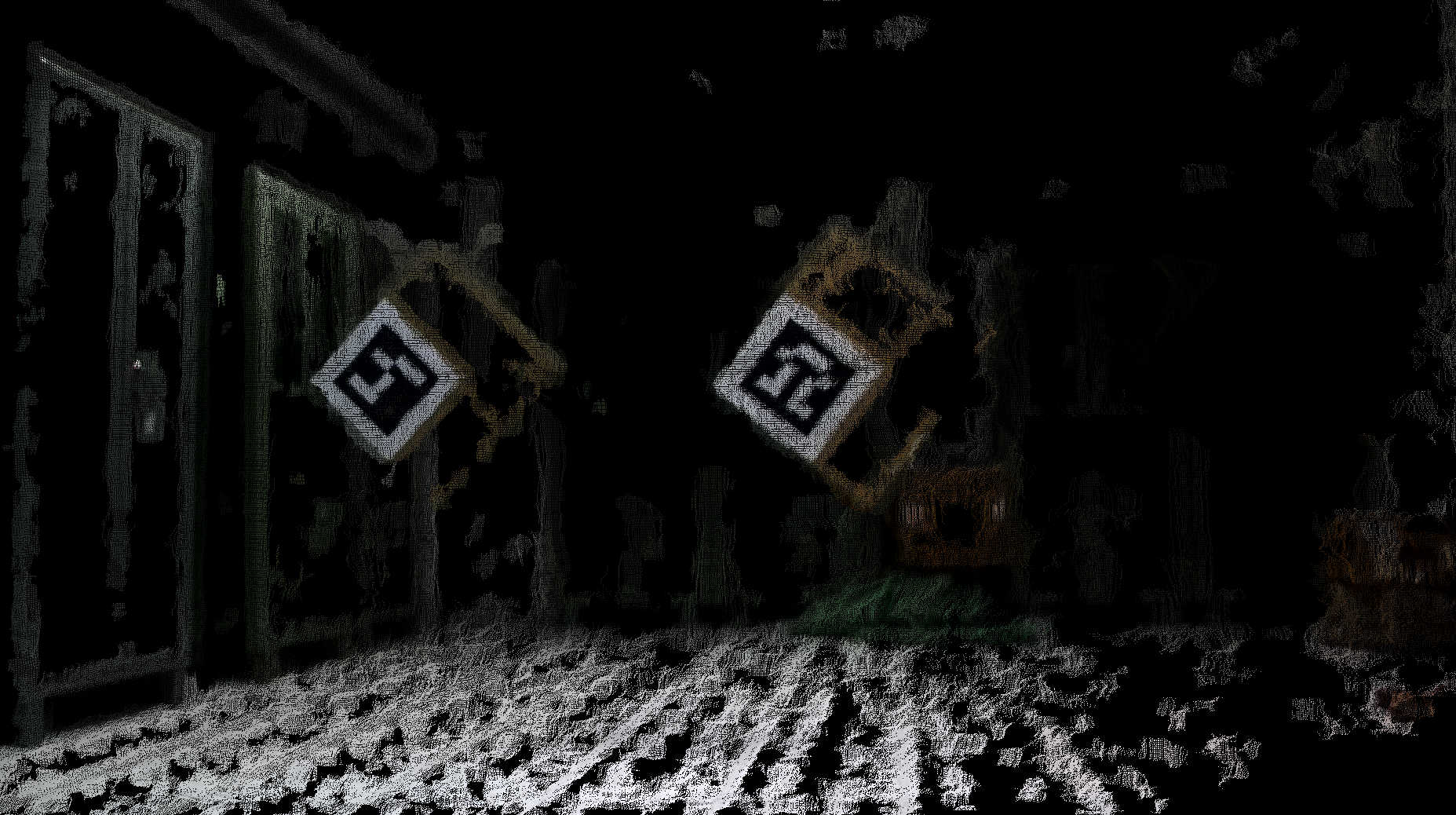} \\

\centering
(a) $[R|t|$ from tape measurement &
\centering
(b) $[R|t|$ from our method \\[6pt]

\end{tabular}

\vspace{-0.4cm}
\caption{Fusion of two pointclouds obtained from two stereo cameras. The pointcloud obtained from camera $C_1$ are transformed to camera $C_2$'s frame-of-reference and then fused. The rigid-body transformation was obtained by the method specified in section \ref{fusing_pcl}. }
\label{fig:manual_machine_1}
\end{figure}

\par
To do this, we introduced the LiDAR which has a 360-degree field-of-view and very precise 3D point co-ordinates can be obtained with it. We use it to find transformations between the cameras. Once, this is done, we can remove the LiDAR; effectively using the LiDAR only for calibrating the cameras.

\par
It is to be noted that the method described in this document calibrates a monocular camera and a LiDAR. If there is a stereo camera, we only calibrate the left camera and the LiDAR. Since, the baseline and stereo camera calibration parameters are already known, calibrating only one of the cameras (left in our case) is sufficient to fuse the point clouds.

\par
If we can find a transformation between two cameras $C_{1}$ and $C_{2}$, we can easily extend the same procedure to obtain transformation between arbitrary number of cameras. Given, two (stereo) cameras, the proposed pipeline finds the transformation that transforms all points in the LiDAR frame to the camera frame.

\par
We first run the algorithm, with $C_{1}$ and LiDAR, $L$, and obtain a $4 \times 4$ matrix,

\begin{equation} \label{eq:LiDAR_to_C1}
T_{LiDAR-to-C_{1}}
\end{equation}

We then run the algorithm, with $C_{2}$ and LiDAR, $L$, and obtain a $4 \times 4$ matrix,

\begin{equation} \label{eq:LiDAR_to_C2}
T_{LiDAR-to-C_{2}}
\end{equation}

Now, to obtain a transform that transforms all points in $C_{1}$ to $C_{2}$, we chain the transforms, $T_{LiDAR-to-C_{1}}$ and $T_{LiDAR-to-C_{2}}$,

\begin{equation} \label{eq:C2_to_C1}
T_{C_{2}-to-C_{1}} = T_{LiDAR-to-C_{1}} \cdot T_{LiDAR-to-C_{2}}^{-1} = T_{LiDAR-to-C_{1}} \cdot T_{C_{2}-to-LiDAR}
\end{equation}

Equation \ref{eq:C2_to_C1}, finds the transform between $C_{2}$ to $C_{1}$, and if these are stereo cameras, we can obtain point clouds and fuse them using this transform. If the transform is very accurate, the two point clouds (from the two stereo cameras) will align properly. However, if there is translation error, when viewing the fused point cloud, hallucinations of objects will be clearly visible, and there will be two of everything. If there is error in the rotation, the points in the two clouds will diverge more and more as the distance from the origin increases.

\par
To verify the method in a more intuitive manner, lidar\_camera\_calibration was used to fuse point clouds obtained from two stereo cameras. We also provide the visualization of the fused point clouds.

\subsection{Manual measurement vs. lidar\_camera\_calibration}
First, we compare the calibration parameters obtained from our method against meticulously measured values using tape by a human.

\begin{figure}[!htbp]
\scriptsize
\centering
\setlength{\tabcolsep}{0.1em}

\begin{tabular}{ p{6cm} p{6cm} }

\includegraphics[width=0.98\linewidth]{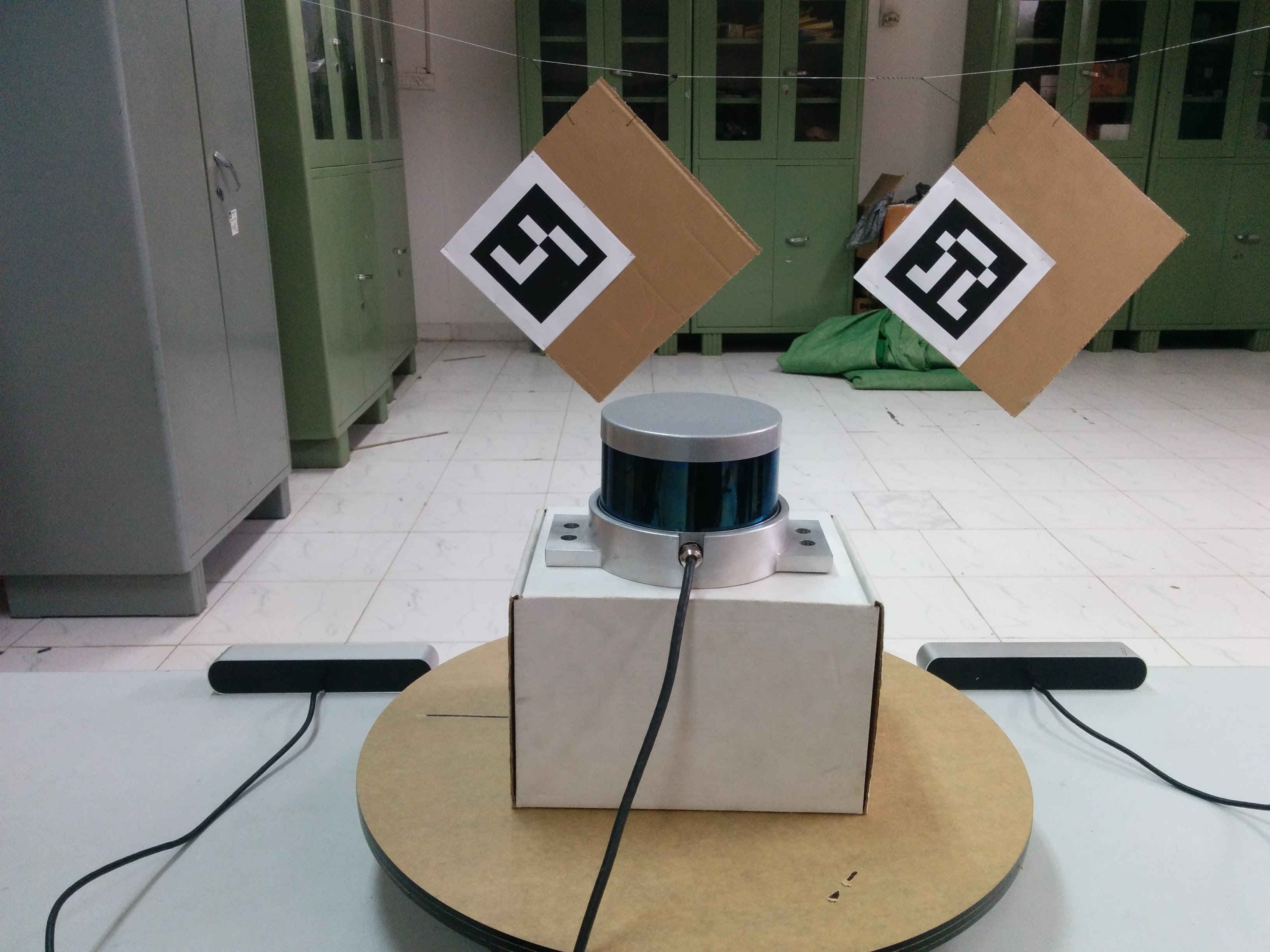} & \includegraphics[width=0.98\linewidth]{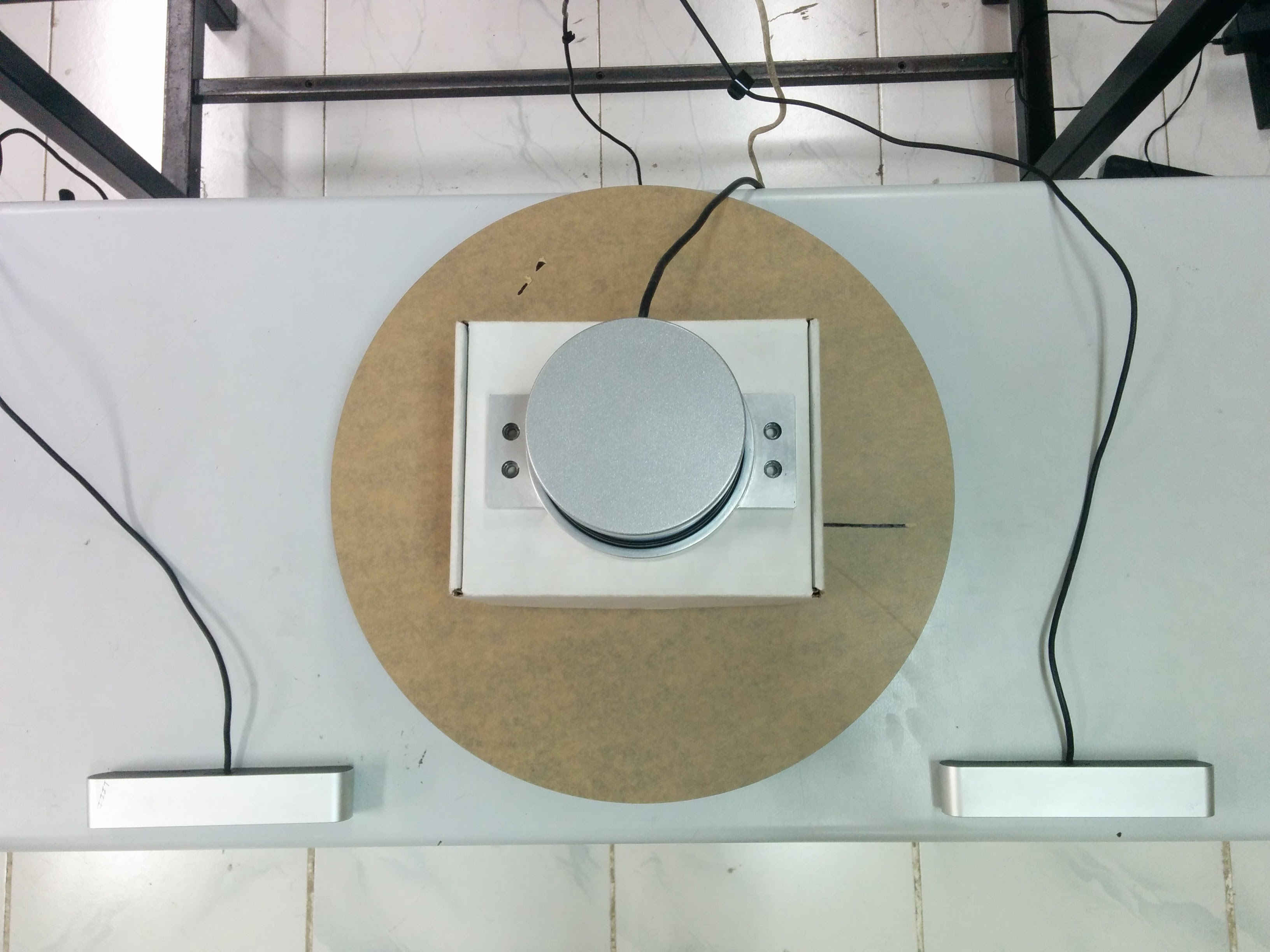}

\end{tabular}

\vspace{-0.4cm}
\caption{Experimental setup for comparing point cloud fusing when using manual measurement versus using transformation obtained from lidar\_camera\_calibration.}
\label{fig:manual_machine_setup}
\end{figure}

The fused point cloud obtained when using manual measurements versus when using the method proposed in this document is shown in the video. Notice the large translation error, even when the two cameras are kept on a planar surface. Hallucinations of markers, cupboards and carton box (in the background) can be seen as a result of the two point clouds not being aligned properly.  

On the other hand, rotation and translation estimated by our package almost perfectly fuses the two individual point clouds. There is a very minute translation error (1-2cm) and almost no rotation error. The fused point cloud is aligned so properly, that one might actually believe that it is a single point cloud, but it actually consists of 2 clouds fused using extrinsic transformation between their sources (the stereo cameras).

The resultant fused point clouds from both manual and lidar\_camera\_calibration methods can be seen on \href{https://youtu.be/AbjRDtHLdz0}{https://youtu.be/AbjRDtHLdz0}.

\subsection{Calibrating cameras kept at 80 degrees}
We also wanted to see the potential of this method and used it to calibrate cameras kept at about 80 degrees and almost no overlapping field-of-view. In principle, with a properly designed experimental setup our method can calibrate cameras with zero overlapping field of view.  

However, to visualize the fusion, we needed a part to be common in both point clouds. We chose a large checkerboard to be seen in both cameras' field-of-view, since it can be used to see how well the point clouds have aligned and if the dimensions of the checkerboard squares are known, one can even estimate the translation errors.

\begin{figure}[!htbp]
\scriptsize
\centering
\setlength{\tabcolsep}{0.1em}

\begin{tabular}{ p{4cm} p{4cm} p{4cm} }

\includegraphics[width=0.98\linewidth]{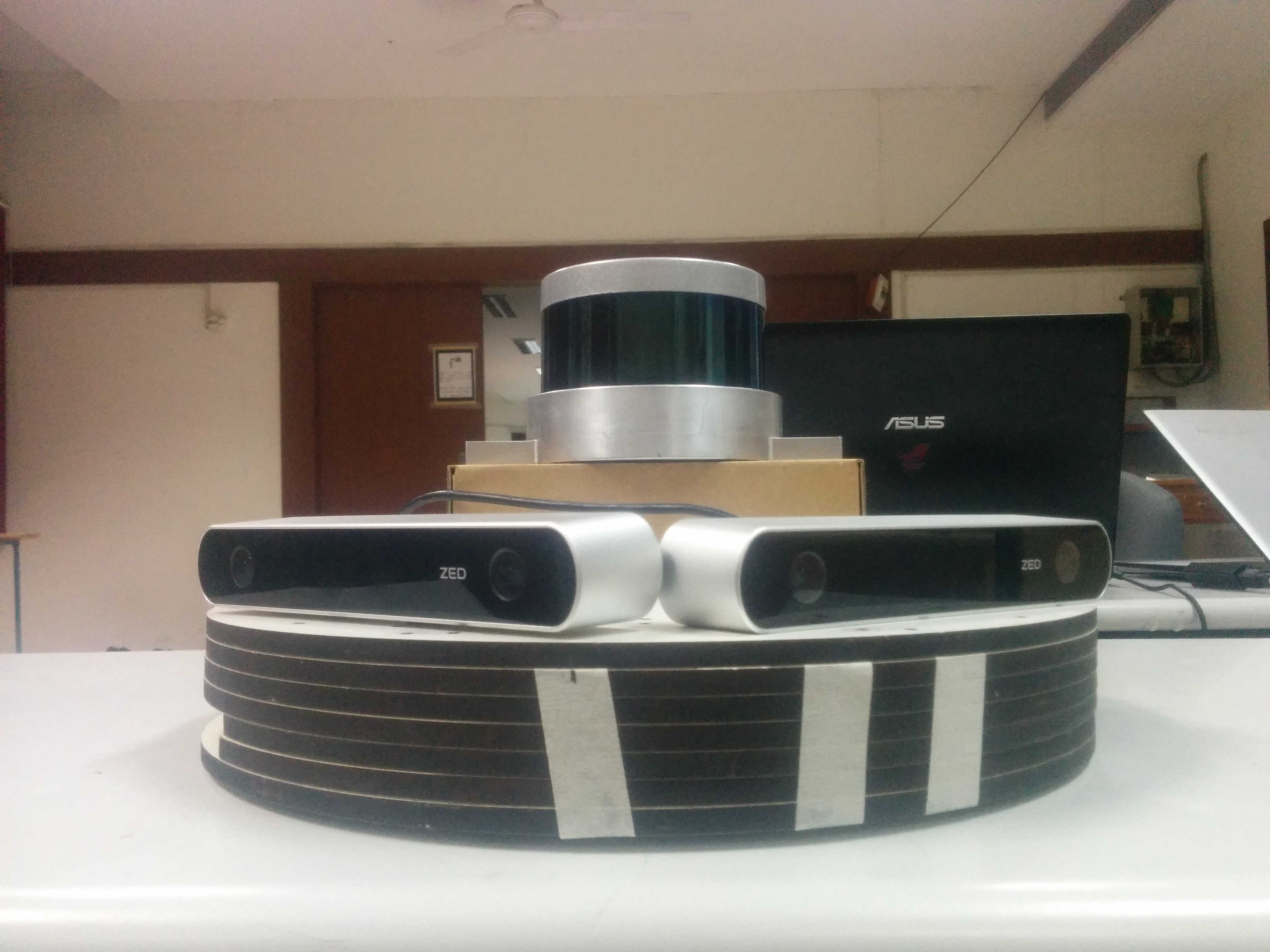} & \includegraphics[width=0.98\linewidth]{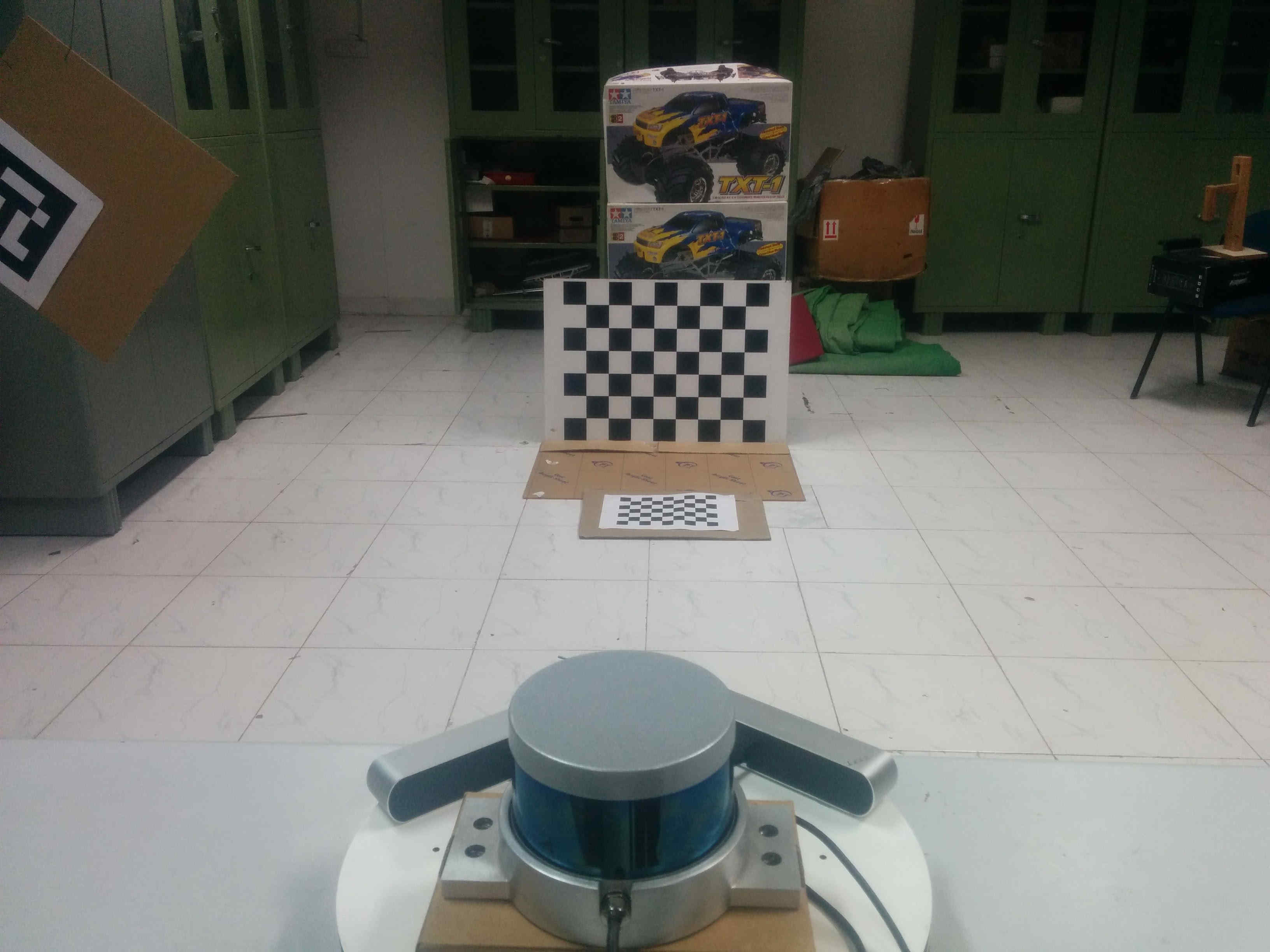} &
\includegraphics[width=0.98\linewidth]{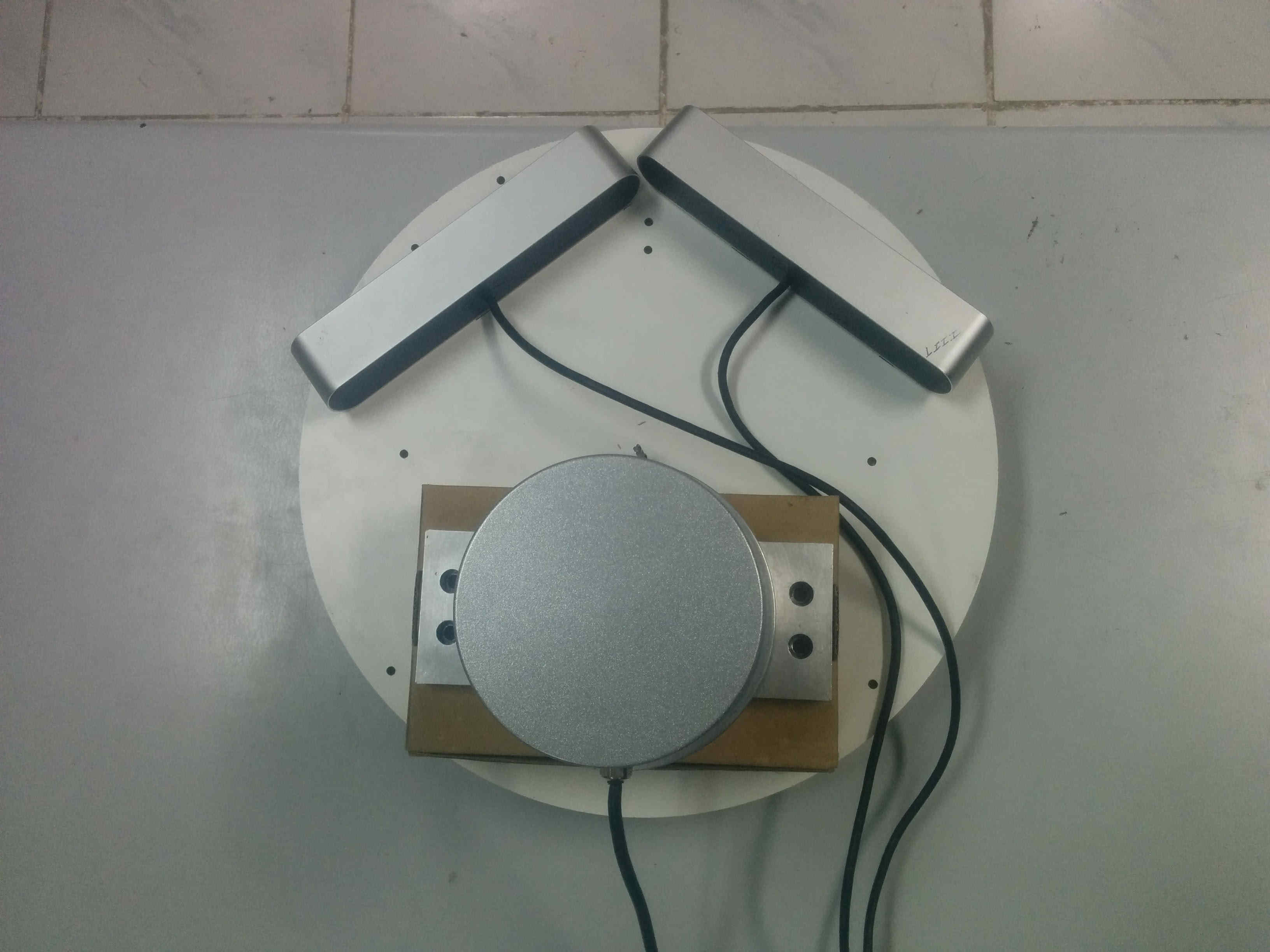}

\end{tabular}

\vspace{-0.4cm}
\caption{Experimental setup with cameras kept at about 80 degrees. The calibration for such configuration shows that our unique method can be, in principle, used for extrinsically calibrating cameras with no field-of-view overlap.}
\label{fig:80deg_Setup}
\end{figure}

There is very less translation error, about 3-4 cm. Also, the ground planes align properly, at all distances, near and far from the camera, implying that the rotations estimated are correct.  

The resultant fused point clouds after extrinsic calibration of stereo cameras kept at approximately 80 degrees using our method can be seen on \\ \href{https://youtu.be/Om1SFPAZ5Lc}{https://youtu.be/Om1SFPAZ5Lc}.

We believe, that better intrinsic calibration of the cameras can help drive down the error to about 1 centimeter or even less.

\section{Results}

As a sanity check, a coarse translation was manually measured using a measuring tape. Rotations however were difficult to measure and their coarse values are omitted. The tabulations below compare the results obtained by using off-the-shelf ICP algorithms and an implementation of the Kabsch algorithm which exploits the information about known correspondences. The Kabsch algorithm repeatedly gives values close to the measurements and the root mean square error (RMSE) is also quite low. Datasets were collected in varying camera and LiDAR configurations with assorted rotations and translations to verify repeatability and accuracy of the proposed method. Separate experiments were performed with a different camera (Point Gray Black Fly) which has a very large focal length as compared to zed stereo camera and the results showed similar accuracy and confirmed robustness of the method proposed.

\begin{table}[!htbp]
\centering
\begin{tabular}{|c|c|c|c|c|}
\hline
Dataset 1 & Tape measurement & $\mu_{offset}$ & ICP & Kabsch \\
\hline
X (in m)       & -0.38 to -0.41 & -0.3837 & -0.4249 & -0.4249 \\
Y (in m)       &  0.04 to -0.06 &  0.0895 &  0.0592 &  0.0591 \\
Z (in m)       &  0.14 to 0.16 &  0.1412 &  0.1399 &  0.1399 \\ \hline
Roll (in deg)  &  unmeasurable         &  -      &  1.5943 &  1.5957 \\
Pitch (in deg) &  unmeasurable         &  -      &  1.4166 &  1.4166 \\
Yaw (in deg)   &  unmeasurable         &  -      & -1.0609 & -1.0635 \\ \hline
RMSE (in m)	 &  -         &  -      &  0.0263 &  0.0240 \\ \hline
\end{tabular}
\caption{$[R|t]$ on dataset 1: manual measurement, mean difference between the two pointclouds, ICP and Kabsch algorithm.}
\end{table}

\begin{table}[!htbp]
\centering
\begin{tabular}{|c|c|c|c|c|}
\hline
Dataset 2 & Tape measurement & $\mu_{offset}$ & ICP & Kabsch \\
\hline
X (in m)       & -0.29 to -0.31 & -0.2725 & -0.3126 & -0.3126 \\
Y (in m)       & -0.25 to -0.27 & -0.0188 & -0.2366 & -0.2367 \\
Z (in m)       &  0.09 to -0.11 &  0.1152 &  0.1195 &  0.1196 \\ \hline
Roll (in deg)  &  unmeasurable         &  -      & -0.3313 & -0.3326 \\
Pitch (in deg) &  unmeasurable         &  -      &  1.6615 &  1.6625 \\
Yaw (in deg)   &  unmeasurable         &  -      & -9.1813 & -9.1854 \\ \hline
RMSE (in m)	 &  -         &  -      &  0.0181 &  0.0225\\ \hline
\end{tabular}
\caption{$[R|t]$ on dataset 2: manual measurement, mean difference between the two pointclouds, ICP and Kabsch algorithm.}
\end{table}

\begin{table}[!htbp]
\centering
\begin{tabular}{|c|c|c|c|c|}
\hline
Dataset 3 & Tape measurement & $\mu_{offset}$ & ICP & Kabsch \\
\hline
X (in m)       & -0.48 to -0.52 & -0.5762 & -0.2562 & -0.5241 \\
Y (in m)       &  0.36 to 0.39 &  0.4583 &  0.1480 &  0.3282 \\
Z (in m)       &  0.07 to 0.09 &  0.1060 &  0.0852 &  0.0565 \\ \hline
Roll (in deg)  &  unmeasurable         &  -      & -6.6721 &  0.4016 \\
Pitch (in deg) &  unmeasurable         &  -      & -2.3090 & -2.2246 \\
Yaw (in deg)   &  unmeasurable         &  -      & -4.5873 & -5.3311 \\ \hline
RMSE (in m)	 &  -         &  -      &  0.2476 &  0.0203 \\ \hline
\end{tabular}
\caption{$[R|t]$ on dataset 3: manual measurement, mean difference between the two pointclouds, ICP and Kabsch algorithm.}
\end{table}

\section{Code and Implementation}
The code is written in C++ and is implemented as a ROS package. It can be found at \href{http://wiki.ros.org/lidar\_camera\_calibration}{http://wiki.ros.org/lidar\_camera\_calibration}. A comprehensive readme file is also available for setting up and getting started with the package is available on the GitHub repo at \href{https://github.com/ankitdhall/lidar\_camera\_calibration}{https://github.com/ankitdhall/lidar\_camera\_calibration}.

\section{Conclusions}

We proposed a novel pipeline to perform accurate LiDAR-camera extrinsic calibration using 3D-3D point correspondences. An experimental setup to find correspondences in each sensor's frame: the camera and the LiDAR. The proposed pipeline uses tags that can be easily printed and stuck on planar surfaces such as cardboards or wooden planks. A point extraction pipeline was implemented to obtain corner points of the cardboards from the pointcloud recorded using the LiDAR. The two sets of point correspondences are used to solve for the $[R|t]$, which gives accurate and repeatable results with different cameras. As opposed to ICP which relies on matching point correspondences, our method, with relatively less number of points and correct correspondences is able to estimate transformation optimally. The method's consistency is further improved by averaging over multiple results.

\par
We also showed how this method could be used to extrinsically calibrate two or more cameras, even when they do not have any overlapping field-of-view. We also successfully demonstrated visually, the quality of the calibration by fusing point clouds and almost perfectly aligning them. An open-source implementation is available in the form of a ROS\cite{ROS} package.

\clearpage


\end{document}